\documentclass[10pt,twocolumn,letterpaper]{article}

\usepackage[pagenumbers]{cvpr} % To force page numbers, e.g. for an arXiv version

\usepackage{graphicx}
\usepackage{amsmath}
\usepackage{amssymb}
\usepackage{booktabs}
\usepackage{enumitem}
\usepackage{placeins}
\usepackage{float}
\usepackage{rotating}

\usepackage{algorithm}
\usepackage{algpseudocode}

\usepackage{multirow}

\usepackage[accsupp]{axessibility}  % Improves PDF readability for those with disabilities.

\usepackage[pagebackref,breaklinks,colorlinks]{hyperref}

\DeclareMathOperator{\up}{\mathsf{up}}
\DeclareMathOperator{\down}{\mathsf{down}}

\usepackage{xcolor}
\usepackage{tabularx}
\newcolumntype{Y}{>{\centering\arraybackslash}X}

\usepackage[capitalize]{cleveref}
\crefname{section}{Sec.}{Secs.}
\Crefname{section}{Section}{Sections}
\Crefname{table}{Table}{Tables}
\crefname{table}{Tab.}{Tabs.}

\def\cvprPaperID{2631} % *** Enter the CVPR Paper ID here
\def\confName{CVPR}
\def\confYear{2023}

\begin{document}

\title{Guided Depth Super-Resolution by Deep Anisotropic Diffusion}

\author{Nando Metzger\thanks{Equal contribution.}\;\quad  Rodrigo Caye Daudt$^\ast$\;\quad  Konrad Schindler\\
\vspace{.0em}
Photogrammetry and Remote Sensing, ETH Zurich\\
{\tt\small \{metzgern,rcayedaudt,schindler\}@ethz.ch}
}

\maketitle

%%%%%%%%% ABSTRACT
\begin{abstract}
    Performing super-resolution of a depth image using the guidance from an RGB image is a problem that concerns several fields, such as robotics, medical imaging, and remote sensing. While deep learning methods have achieved good results in this problem, recent work highlighted the value of combining modern methods with more formal frameworks. In this work, we propose a novel approach which combines guided anisotropic diffusion with a deep convolutional network and advances the state of the art for guided depth super-resolution. The edge transferring/enhancing properties of the diffusion are boosted by the contextual reasoning capabilities of modern networks, and a strict adjustment step guarantees perfect adherence to the source image. We achieve unprecedented results in three commonly used benchmarks for guided depth super-resolution. The performance gain compared to other methods is the largest at larger scales, such as $\times$32 scaling. Code\footnote{{ \url{https://github.com/prs-eth/Diffusion-Super-Resolution}}} for the proposed method is available to promote reproducibility of our results.
\end{abstract}

%%%%%%%%% BODY TEXT
\section{Introduction}

It is a primordial need for visual data analysis to increase the resolution of images after they have been captured. In many fields one is faced with images that, for technical reasons, have too low resolutions for the intended purposes, \eg, MRI scans in medical imaging~\cite{zhang2018longitudinally}, multi-spectral satellite images in Earth observation~\cite{lanaras2018super}, thermal surveillance images \cite{almasri2018rgb} and depth images in robotics \cite{eichhardt2017image}.
In some cases, an image of much higher resolution is available in a different imaging modality, which can act as a \emph{guide} for super-resolving the low-resolution \emph{source} image, by injecting the missing high-frequency content. 
For instance, in Earth observation, the guide is often a panchromatic image (hence the term "pan-sharpening"), whereas in robotics a conventional RGB image is often attached to the same platform as a TOF camera or laser scanner.
In this paper, we focus on super-resolving depth images guided by RGB images, but the proposed framework is generic and can be adapted to other sensor combinations, too.

\begin{figure}[t]
    \centering
    \includegraphics[width=0.475\textwidth]{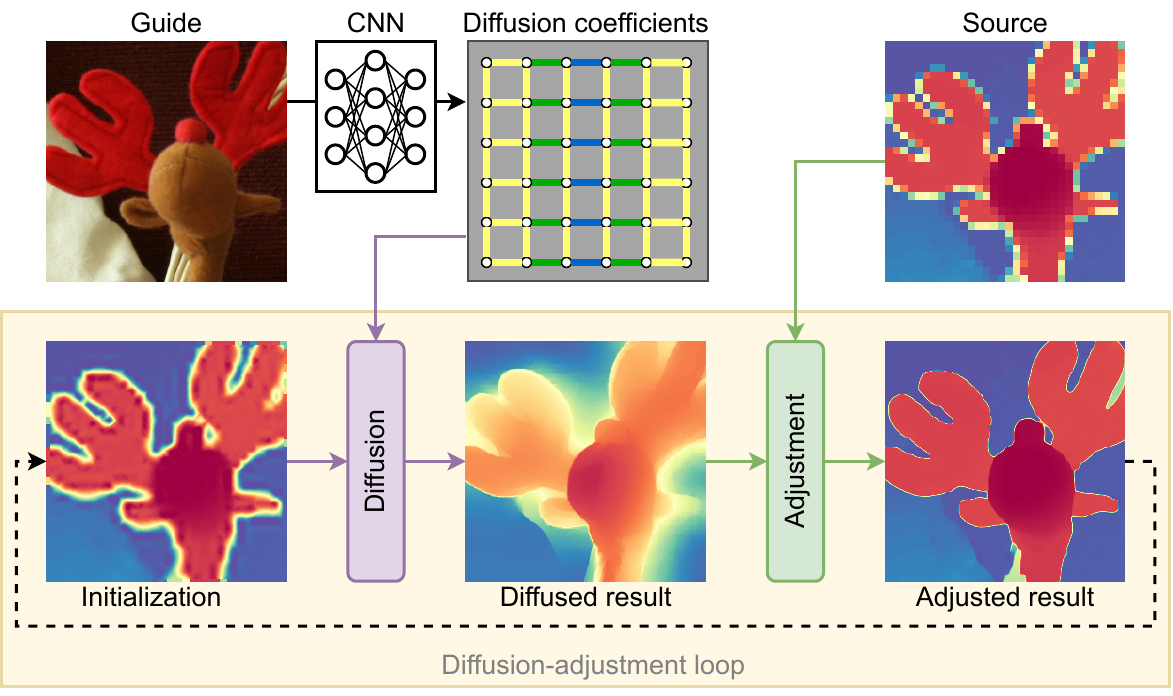}
    \caption{We super-resolve a low-resolution depth image by finding the equilibrium state of a constrained anisotropic diffusion process.
    Learned diffusion coefficients favor smooth depth within objects and suppress diffusion across discontinuities. They are derived from the guide with a neural feature extractor that is trained by back-propagating through the diffusion process. }
    \vspace{-0.3cm}
    \label{fig:teaser}
\end{figure}

Research into guided super-resolution has a long history \cite{patterson1992,izraelevitz1994}. The proposed solutions range from classical, entirely hand-crafted schemes \cite{ham2017robust} to fully learning-based methods \cite{Tak-Wai2016}, while some recent works have combined the two schools of thought, with promising results~\cite{riegler2016deep, de2022learning}. 
Many classical methods boil down to an image-specific optimization problem that must be solved at inference time, which often makes them slow and memory-hungry. Moreover, they are limited to low-level image properties of the guide, such as color and contrast, and lack the high-level image understanding and contextual reasoning of modern neural networks. On the positive side, by design, they can not overfit the peculiarities of a training set and tend to generalize better.
Recent work on guided super-resolution has focused on deep neural networks. Their superior ability to capture latent image structure has greatly advanced the state of the art over traditional, learning-free approaches. Still, these learning-based methods tend to struggle with sharp discontinuities and often produce blurry edges in the super-resolved depth maps. Moreover, like many deep learning systems, they degrade -- often substantially -- when applied to images with even slightly different characteristics.
Note also that standard feed-forward architectures cannot guarantee a consistent solution: feeding the source and guiding images through an encoder-decoder structure to obtain the super-resolved target will, by itself, not ensure that downsampling the target will reproduce the source.

We propose a novel approach for guided depth super-resolution which combines the strengths of optimization-based and deep learning-based super-resolution.
In short, our method is a combination of anisotropic diffusion (based on the discretized version of the heat equation) with deep feature learning (based on a convolutional backbone). 
The diffusion part resembles classical optimization approaches, solved via an iterative diffusion-adjustment loop. Every iteration consists of (1) an anisotropic diffusion step~\cite{perona1990scale, liu2013guided, daudt2019guided, daudt2021weakly}, with diffusion weights driven by the guide in such a way that diffusion (i.e., smoothing) is low across high-contrast boundaries and high within homogeneous regions; and (2) an adjustment step that rescales the depth values such that they exactly match the low-resolution source when downsampled.
To harness the unmatched ability of deep learning to extract informative image features, the diffusion weights are not computed from raw brightness values but are set by passing the guide through a (fully) convolutional feature extractor.
An overview of the method is depicted in \cref{fig:teaser}.
The technical core of our method is the insight that such a feature extractor can be trained end-to-end to optimally fulfill the requirements of the subsequent optimization, by back-propagating gradients through the iterations of the diffusion loop.
Despite its apparent simplicity, this hybrid approach delivers excellent super-resolution results. In our experiments, it consistently outperforms prior art on three different datasets, across a range of upsampling factors from $\times$4 to $\times$32. 
In our experiments, we compare it to six recent learning methods as well as five different learning-free methods. For completeness, we also include a learning-free version of our diffusion-adjustment scheme and show that it outperforms all other learning-free methods.
Beyond the empirical performance, our method inherits several attractive properties from its ingredients: the diffusion-based optimization scheme ensures strict adherence to the depth values of the source, crisp edges, and a degree of interpretability; whereas deep learning equips the very local diffusion weights with large-scale context information, and offers a tractable, constant memory footprint at inference time.
In summary, our contributions are:
\begin{enumerate}[parsep=2pt]
    \item We develop a hybrid framework for guided super-resolution that combines deep feature learning and anisotropic diffusion in an integrated, end-to-end trainable pipeline;
    \item We provide an implementation of that scheme with constant memory demands, and with inference time that is constant for a given upsampling factor and scales linearly with the number of iterations;
    \item We set a new state of the art for the Middlebury~\cite{Scharstein2001}, NYUv2~\cite{Shelhamer2017} and DIML~\cite{DIML1} datasets, for upsampling factors from 4$\times$ to 32$\times$, and provide empirical evidence that our method indeed guarantees exact consistency with the source image.
\end{enumerate}

\section{Related Work} \label{sec:related}

\paragraph{Learning-free methods.}
Early work on guided super-resolution consisted mostly of optimization methods. Several such methods~\cite{diebel2005application, xie2015edge, mac2012patch} employ random field models to solve this problem. 
Other traditional methods rely on filters, such as the bilateral filter \cite{yang2007spatial}, the guided filter~\cite{he2012guided}, the weighted median filter~\cite{ma2013constant}, the weighted mode filter~\cite{min2011depth}, or the Static-Dynamic filter~\cite{ham2017robust}. Liu \etal~\cite{liu2013guided} showed an early, learning-free application of anisotropic diffusion for guided depth enhancement. De Lutio \etal~\cite{lutio2019guided} fit a pixel-wise MLP to map the guide to the target and obtain excellent results when compared to other learning-free methods, whereas Uezato \etal~\cite{uezato2020guided} adapt Deep Image Prior~\cite{ulyanov2018deep} to fuse the guide and the source images.

\paragraph{Learning-based methods.}
Other avenues have also been explored, such as the auto-regressive model proposed in \cite{wang2019multi}.
More recent architectures \cite{song2020channel,Ye2020} have used successfully applied transformer modules to this problem. With the advent of deep learning in the past decade, several authors have proposed feedforward networks for upsampling depth images. \textit{MSG-Net}~\cite{Tak-Wai2016} is a U-Net shaped architecture that embeds the source at its lowest scale and learns the residual errors of bicubic interpolations. Kim \etal~\cite{Kim2021} propose Deformable Kernel Networks (\textit{DKN} and its efficient implementation \textit{FDKN}), that learn sparse and spatially-invariant filter kernels. He \etal~\cite{he2021} employ a high-frequency guidance module to embed the guide details into the depth map. Wen \etal~\cite{wen2018deep} use convolutional kernels of different sizes while also using data fidelity as a convergence criterion of their iterative refinement.

\paragraph{Hybrid methods.}
A final group of methods applies deep learning methods within formal frameworks that constrain their solution and improve the methods' inductive biases for guided super-resolution.
Riegler \etal~\cite{riegler2016deep} unroll the optimization steps of a first-order primal-dual algorithm into a neural network, such that they can train their deep feature extractor in an end-to-end manner. De Lutio \etal~\cite{de2022learning} use the implicit function theorem to propagate through a graph-based, MRF-style optimizer, combining deep learning within a framework inspired by traditional approaches~\cite{diebel2005application,park2019planar,rossi2020joint}. Work exploring this last family of algorithms has reported promising results from combining the contextual reasoning capabilities of convolutional neural networks (CNNs) with the explicit constraints of formal frameworks. Our work can also be regarded as a member of this group of methods.

\section{Method}

\begin{figure*}[t]
    \centering
    \includegraphics[width=\textwidth]{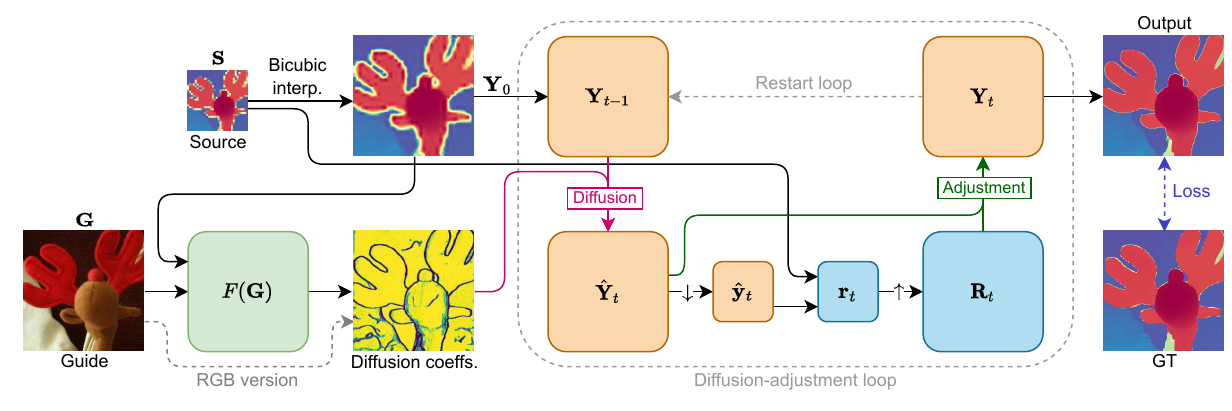}
    \vspace{-0.6cm}\caption{  The diffusion step performs anisotropic diffusion on the depth image using diffusion weights given by a CNN. The adjustment step ensures that the diffused result matches the source image at its original resolution. Gradients from the loss function can be back-propagated back to the CNN, making the system end-to-end trainable.}
    \vspace{-0.3cm}
    \label{fig:main_schematic}
\end{figure*}

We first describe the diffusion-adjustment framework used for guided super-resolution, then we explain how deep learning is integrated into that framework. The latter part allows us to exploit high-level context in the guide image to find optimal coefficients for the diffusion operator.

\subsection{Diffusion-adjustment}

Anisotropic diffusion~\cite{perona1990scale} is a form of iterative edge-aware filtering initially proposed with the aim of performing intra-region smoothing while avoiding inter-region smoothing. The filtering is done in a way analog to solving a discretized heat equation with anisotropic (\ie, spatially varying) diffusion weights. The idea of computing these diffusion weights from a separate (coregistered) guide image has been explored in the context of edge enhancement~\cite{liu2013guided} and semantic segmentation~\cite{daudt2019guided, daudt2021weakly}.

We are given a source image $\mathbf{S} \in \mathbb{R}^{\frac{H}{s} \times \frac{W}{s}}$ and a guide $\mathbf{G} \in \mathbb{R}^{H \times W \times C}$, where $C=3$ for RGB images or a larger number for deep features. The first step in our approach is to initialize $\mathbf{Y}_0 \in \mathbb{R}^{H \times W}$ with an upsampled version of $\mathbf{S}$. As we will show later, the exact initialization of $\mathbf{Y}_0$ has little impact on the final result. We can then define a diffusion step as:
\begin{equation}
    \hat{\mathbf{y}}^{p}_t = \mathbf{y}^p_{t-1} + \lambda \cdot \!\!\sum_{n \in \mathcal{N}^p_4} ( \mathbf{y}^n_{t-1} - \mathbf{y}^p_{t-1} ) \cdot c(\mathbf{g}^p, \mathbf{g}^n) \;,
\end{equation}
where $\mathbf{y}^p_t$ denotes the pixel value of $\mathbf{Y}_t$ at location $p$ (and similarly for $\mathbf{g}^p$). $\mathcal{N}^p_4$ denotes the 4-neighborhood of pixel $p$. Note that this construction connects all pixels in the image into a (planar) graph. $\lambda$ is a strictly positive hyperparameter that regulates the update and ensures stability. When using $\mathcal{N}^p_4$ connectivity it should be set to $\lambda < \frac{1}{4}$. The function $c: (\mathbb{R}^C, \mathbb{R}^C) \to (0,1)$ produces diffusion coefficients for neighboring pairs of pixels based on their values in the guide. Following \cite{perona1990scale}, we use
\begin{equation}\label{eq:c}
c(\mathbf{g}^p, \mathbf{g}^n)=\frac{\kappa^2}{\kappa^2+\| \mathbf{g}^p - \mathbf{g}^n\|_2^2}
\end{equation}
where $\kappa$ is a hyperparameter that regulates the sensitivity to gradients in $\mathbf{G}$. Note that $c$ is symmetrical, $c(\mathbf{g}^p, \mathbf{g}^n) = c(\mathbf{g}^n, \mathbf{g}^p)$. Traditional (non-guided) anisotropic diffusion is a special case of the formulation above where $\mathbf{G} = \mathbf{Y}_{t-1}$.

When applied to a single image, anisotropic diffusion has edge-enhancing properties~\cite{perona1990scale}. In guided diffusion, where the diffusion weights are computed from a separate guide image, the diffusion process transfers edges from the guide to the target image~\cite{liu2013guided, daudt2019guided, daudt2021weakly}. This is the property that motivates its use in our guided depth super-resolution framework: the diffusion allows us to precisely recover depth discontinuities in the upsampling result.

By itself, diffusion does not take into account the constraint provided by the source $\mathbf{S}$. \Ie, with the machinery introduced so far $\mathbf{Y}_t$ would approach a constant image with pixel values $\mu (\mathbf{Y}_0)$ as $t \rightarrow \infty$, losing all information. To tie the output to the source image $\mathbf{S}$, every diffusion step is followed by an adjustment step that restores compatibility with $\mathbf{S}$. Doing so guarantees that the output of every iteration, and therefore also the equilibrium $\lim_{t \rightarrow \infty} \mathbf{Y}_t$, is a valid upsampling of $\mathbf{S}$.

The adjustment is done by simply re-scaling patches of $\hat{\mathbf{Y}}_t$ such that, when downsampled to the source resolution, they exactly match $\mathbf{S}$. More formally, the adjustment step can be written as
\vspace{-0.6em}
\begin{equation}
    \mathbf{Y}_{t} = \hat{\mathbf{Y}}_{t} \cdot \underbrace{\up \Bigg( \overbrace{ \frac{\mathbf{S}}{\down(\hat{\mathbf{Y}}_{t})}}^{\mathbf{r}_{t}} \Bigg)}_{\mathbf{R}_{t}},
    \vspace{-0.1cm}
\end{equation}
where $\down$ denotes a linear downsampling operator, and $\up$ denotes nearest neighbor upsampling. $\mathbf{r}_{t}$ and $\mathbf{R}_{t}$ are the adjustment ratios at different scales. After this adjustment step, it is guaranteed that the target matches the source at a lower scale, \ie, $\down ( \mathbf{Y}_t ) = \mathbf{S}$.

\begin{figure}[t]
    \centering
    \includegraphics[width=0.4\textwidth]{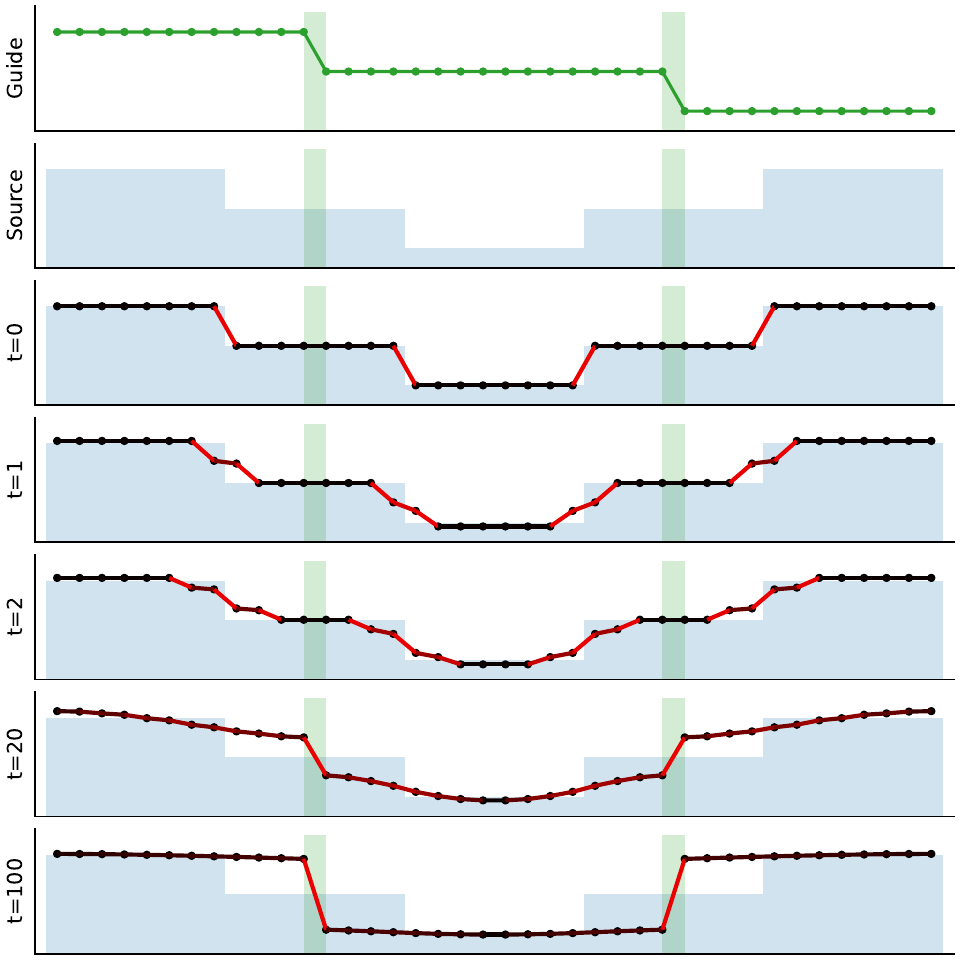}
    \caption{Diffusion diminishes gradients in the target signal (highlighted in red). Strong gradients in the guide signal (marked in green) impede diffusion and are thus transferred to the target. Additionally, the target is constrained to adhere to the low-resolution source (depicted in blue).}
    \label{fig:1d_diffusion_ex_v2}
    \vspace{-0.3cm}
\end{figure}

An illustration of the method is displayed in \cref{fig:main_schematic}. Furthermore, we show the evolution during the diffusion process for a 1D example in \cref{fig:1d_diffusion_ex_v2}. Gradients in the diffused signal quickly dissipate where gradient in the guide are low, but diffusion (almost) stops at edges of the guide. Consistency with the source $\mathbf{S}$ is preserved throughout the process.

\subsection{Deep Feature-Guided Anisotropic Diffusion}

Recent work has shown that the context features extracted by with CNNs over large receptive fields can massively boost super-resolution based on low-order graphs~\cite{de2022learning}. A main message of our paper is that this idea is even more powerful when combined with diffusion on the graph. 

Let $F: \mathbb{R}^{H \times W \times C_1} \to \mathbb{R}^{H \times W \times C_2}$ be a neural feature extractor. In our experiments, $F$ is a U-Net~\cite{ronneberger2015u} with ResNet-50~\cite{he2016deep} backbone pretrained on ImageNet \cite{deng2009imagenet} and $C_2 = 64$, but any other neural architecture could be used, as long as the spatial dimensions of the output match those of the input. While $F$ could be applied directly to the RGB guide $\mathbf{G}$, we concatenate the upsampled source $\up (\mathbf{S})$ as a fourth channel to support object separation based on coarse depth cues, so $C_1 = 4$.

Previous work that connected guided anisotropic diffusion with deep learning~\cite{daudt2019guided, daudt2021weakly} was restricted to using it only as a post-processing step at inference time, due to excessively high memory requirements, and a risk of vanishing/exploding gradients caused by the iterative nature of the algorithm. We found that it is in fact possible to propagate gradients through the diffusion process with the following scheme:
\begin{enumerate}[parsep=2pt]
    \item We compute the $c(\mathbf{g}^p, \mathbf{g}^n)$ only once before diffusion starts and freeze them. This is in contrast to earlier attempts that diffused the guide $\mathbf{G}$ alongside $\mathbf{Y}$ and had to update the $c(\mathbf{g}^p, \mathbf{g}^n)$ at every iteration \cite{daudt2019guided, daudt2021weakly};
    \item We back-propagate gradients only through the last $N_{\text{grad}}$ diffusion-adjustment iterations (in practice $N_\text{grad}\approx10^3$). \Ie, during training the feature extractor receives training signals from the later stages of the diffusion process.
\end{enumerate}
With these modifications, we can effectively back-propagate gradients all the way to the feature extractor $F$ and train the entire pipeline end-to-end. Furthermore, note that $\kappa$ in \cref{eq:c} is also trainable, and therefore does not need to be chosen manually.

At training time we use a random number of iterations without gradient before computing iterations with gradient. We define $N_{\text{pre}}$ as the maximum number of iterations without gradient updates at training time and $N_{\text{grad}}$ as the number of iterations with gradients. We found that this randomization helps to speed up the convergence at inference time (see supplementary material).
For inference we use a constant $N = N_{\text{pre}} + N_{\text{grad}}$ iterations.
The rationale behind using the last $N_{\text{grad}}$ iterations for the computation of gradients is that we are interested in the steady-state equilibrium of the system, and therefore need a high number of iterations. We study the effect of $N_{\text{pre}}$ and $N_{\text{grad}}$ and the associated computational costs in \Cref{sec:experiments}, and find that there is little gain for $N_{\text{grad}} > 1024$, a setting at which the system can still be trained on a single GPU. Finally, we point out that while the algorithm needs a relatively large number of iterations (we find 8000 to be a suitable number), the diffusion and adjustment operators are extremely lightweight and parallelizable, and as a result, our algorithm is still faster than other optimization approaches.

\section{Experiments}\label{sec:experiments}

We evaluate our method on three different RGB-D datasets commonly used for super-resolution, namely Middlebury 2005-2014 \cite{Scharstein2007,Scharstein2001,Scharstein2014,Scharstein2003,Hirschmuller2007}, NYUv2 \cite{Shelhamer2017},  and DIML \cite{DIML1,DIML2,DIML3,DIML4}. We closely follow the setup of \cite{de2022learning}, including the train/validation/test splits, data prepossessing and usage of evaluation metrics.

\textbf{Middlebury} \cite{Scharstein2007, Scharstein2001, Scharstein2003,  Scharstein2014, Hirschmuller2007} consists of 50 photogrammetrically created high-resolution depth maps and their associated RGB images from the years from 2005 to 2014. Five samples are reserved for validation and testing each, while the rest is used for training. The depth maps contain data gaps, which are also present in the source images.
This dataset provides the most accurate ground truth among the considered datasets.

\textbf{NYUv2} \cite{Shelhamer2017} was captured with a Microsoft Kinect depth camera. It consists of a total of 1449 RGB-D images, from which 849 are used for training, and 300 each for validation and testing.

\textbf{DIML} \cite{DIML1,DIML2,DIML3,DIML4} contains 2 million RGB-D samples, from which we only use the high quality indoor samples that were acquired with a Microsoft Kinect depth camera. 1440 are training samples, 169 are validation samples and 503 are test samples.

\begin{table}[t]
    \centering
    \resizebox{0.475\textwidth}{!}{%
    \begin{tabular}{l | cccc}
    % \toprule
     & $\times4$  & $\times8$  & $\times16$  & $\times32$  \\
     & MSE / MAE & MSE / MAE & MSE / MAE & MSE / MAE  \\ \midrule \midrule
    \multicolumn{5}{c}{\textbf{Middlebury}}  \\ \midrule
    MSG & 4.13 / 0.22 & 10.5 / 0.43 & 34.2 / 1.06 & 92.9 / 2.55 \\
    DKN & 4.29 / 0.18 & 11.2 / 0.38 & 47.6 / 1.42 & 119 / 3.35 \\
    FDKN & 3.60 / 0.16 & 10.4 / 0.37 &  38.5 / 1.18 & 112 / 3.23 \\
    PMBA & 4.72 / 0.25 & 9.48 / 0.38 & 30.6 / 0.89 & 175 / 5.15 \\
    FDSR & 7.72 / 0.35 & 23.2 / 0.69 & 55.4 / 1.51 & 179 / 4.11 \\
    LGR & 3.04 / 0.13 & 7.26 / 0.24 & 24.7 / 0.67 & 63.4 / 1.75 \\
    \textbf{DADA} & \textbf{2.52} / \textbf{0.11} & \textbf{5.63} / \textbf{0.20} & \textbf{15.6} / \textbf{0.47} & \textbf{47.6} / \textbf{1.35} \\ %\bottomrule

    \midrule \midrule 

     \multicolumn{5}{c}{\textbf{NYUv2}}  \\ \midrule
    MSG & 6.85 / 0.81 & 24.1 / 1.66 &  84.5 / 3.35 & 262 / 6.70 \\
    DKN & 11.4 / 1.03 &  29.8 / 1.82 & 115 / 4.01 & 364 / 8.33 \\
    FDKN & 8.07 / 0.85 &  29.9 / 1.80 & 113 / 3.95 & 365 / 8.39 \\
    PMBA & 10.8 / 0.93 & 31.5 / 1.79 & 84.9 / 3.26 & 449 / 9.11 \\
    FDSR & 10.5 / 0.94 & 35.4 / 1.96 & 179 / 4.68 & 771 / 11.8 \\
    LGR & 6.45 / 0.73 & 19.6 / 1.42 & 67.5 / 2.90 & 253 / 6.50 \\
    \textbf{DADA} & \textbf{4.87} / \textbf{0.64} & \textbf{17.1} / \textbf{1.33} & \textbf{59.2} / \textbf{2.65} & \textbf{223} / \textbf{5.76} \\ %\bottomrule

    \midrule \midrule 

    \multicolumn{5}{c}{\textbf{DIML}}  \\ \midrule
    MSG & 1.73 / 0.22 & 4.13 / 0.40 & 13.0 / 0.93 & 55.0 / 2.56 \\
    DKN & 3.47 / 0.33 & 5.47 / 0.45 & 19.3 / 1.20 & 91.5 / 3.75 \\
    FDKN & 2.2 / 0.23 & 5.95 / 0.47 & 20.8 / 1.24 & 89.9 / 3.75 \\
    PMBA & 3.05 / 0.31 & 5.87 / 0.47 & 13.8 / 0.87 & 55.1 / 2.30 \\
    FDSR & 2.75 / 0.29 & 8.40 / 0.66 & 32.9 / 1.66 & 124 / 4.39 \\
    LGR & 1.68 / 0.20 & 3.51 / 0.31 & 9.45 / 0.68 & 57.0 / 2.47 \\
    \textbf{DADA} & \textbf{1.30} / \textbf{0.17} & \textbf{2.87} / \textbf{0.27} & \textbf{7.75} / \textbf{0.60} & \textbf{38.6} / \textbf{1.90} \\ %\bottomrule
    \end{tabular}%
    }
    \caption{Performance comparison of learning-based methods in terms of MSE (in cm$^2$) and MAE (in cm). DADA consistently outperforms all other methods, especially at large scaling factors. We report the variability of our method in the supplementary material.}
    \vspace{-0.3cm}
    \label{tab:cmp_l}
\end{table}

\begin{table}[t]
    \centering
    \resizebox{0.475\textwidth}{!}{%
    \begin{tabular}{l | cccc}
    % \toprule
     & $\times4$  & $\times8$  & $\times16$  & $\times32$  \\
     & MSE / MAE & MSE / MAE & MSE / MAE & MSE / MAE  \\ \midrule \midrule
    \multicolumn{5}{c}{\textbf{Middlebury}}  \\ \midrule
    Bicubic & 13.3 / 0.55 & 30.0 / 1.10 & 68.5 / 2.13 & 143 / 3.87 \\
    GF & 33.3 / 1.27 & 40.5 / 1.49 & 67.4 / 2.21 & 134 / 3.82  \\
    SD & 24.9 / 0.46 & 82.5 / 0.86 & 511 / 1.73 & 4062 / 3.37 \\
    PixT & 39.8 / 0.79 & 32.7 / 0.82 &  41.5 / \textbf{1.24} & 107 / 2.71 \\
    LGR$^\dag$ & 14.8 / 0.42 & 68.3 / 0.83 & 297 / 1.69 & 897 / 3.31 \\
    \textbf{DADA}$^\dag$  & \textbf{11.1} / \textbf{0.40} & \textbf{18.9} / \textbf{0.70} & \textbf{36.7} / 1.30 & \textbf{84.1} / \textbf{2.58} \\ 

    \midrule \midrule 

    \multicolumn{5}{c}{\textbf{NYUv2}}  \\ \midrule
    %  &  & \multicolumn{2}{c}{x4}  & \multicolumn{2}{c}{x8}  & \multicolumn{2}{c}{x16}  & \multicolumn{2}{c}{x32}  \\
    %  &  & MSE & MAE & MSE & MAE & MSE & MAE & MSE & MAE  \\ \midrule \midrule
    Bicubic &  31.9 / 1.59  & 89.9 / 3.17  & 242 / 6.01 & 588 / 10.5 \\
    GF &  114 / 3.91 & 142 / 4.47 & 249 / 6.34 & 556 / 10.4 \\
    SD & 36.0 / 1.31 & 105 / 2.57 & 533 / 5.07 & 3246 / 10.4 \\
    PixT & 112 / 3.61 & 122 / 3.86 & 219 / 5.40 & 759 / 11.6 \\
    LGR$^\dag$ &  19.0 / 1.11 &  68.4 / 2.30 & 264 / 4.56 & 790 / 9.31 \\
    \textbf{DADA}$^\dag$  & \textbf{18.1} / \textbf{1.05} & \textbf{49.9} / \textbf{2.05} & \textbf{125} / \textbf{3.88} & \textbf{328} / \textbf{7.50} \\ 

    \midrule \midrule 

    \multicolumn{5}{c}{\textbf{DIML}}  \\ \midrule
    %  &  & \multicolumn{2}{c}{x4}  & \multicolumn{2}{c}{x8}  & \multicolumn{2}{c}{x16}  & \multicolumn{2}{c}{x32}  \\
    %  &  & MSE & MAE & MSE & MAE & MSE & MAE & MSE & MAE  \\ \midrule \midrule
    Bicubic & 10.4 / 0.63 & 28.6 / 1.32 & 73.2 / 2.68 & 187 / 5.37 \\
    GF & 25.6 / 1.45 & 34.1 / 1.77 & 66.3 / 2.74 & 165 / 5.18 \\
    SD & 10.5 / 0.40 & 44.9 / 0.83 & 411 / 1.91 & 5905  /  4.45 \\
    PixT &  20.7 / 1.15 & 23.0 / 1.26 & 39.3 / 1.78 & 141 / 4.19 \\
    LGR$^\dag$ & 7.02 / 0.35 & 15.2 / 0.67 & 133 / 1.72 & 815 / 3.98 \\
    \textbf{DADA}$^\dag$  &  \textbf{4.40} / \textbf{0.28} & \textbf{9.35} / \textbf{0.51} & \textbf{21.3} / \textbf{1.15} & \textbf{72.6} / \textbf{3.02} \\ 
    \end{tabular}%
    }
    \caption{Performance comparison of learning-free methods in terms of MSE (in cm$^2$) and MAE (in cm). Our diffusion-adjustment scheme, without feature learning, outperforms other learning-free methods. $^\dag$ marks learning-free variants of hybrid methods.}
    \vspace{-0.3cm}
    \label{tab:cmp_nl}
\end{table}

\subsection{Experimental Setup}

We compare our deep anisotropic diffusion-adjustment network (\emph{DADA}) against a broad, representative range of guided super-resolution methods, both learning-based and learning-free. We also include simple non-guided bicubic upsampling (\emph{Bicubic}) \cite{keys1981cubic} as a baseline and sanity check.
The learning-based methods we consider are MSG-Net (\emph{MSG})~\cite{Tak-Wai2016}, Deformable Kernel Network (\emph{DKN})~\cite{Kim2021}, Fast Deformable Kernel Network (\emph{FDKN})~\cite{Kim2021},  Fast Depth Super-Resolution (\emph{FDSR})~\cite{he2021}, PMBANet (\emph{PMBA})~\cite{Ye2020} and Learned Graph Regularizer (\emph{LGR})~\cite{de2022learning}. 
Learning-free methods in our evaluation are the guided filtering (\emph{GF})~\cite{he2012guided}, Static/Dynamic filtering (\emph{SD})~\cite{ham2017robust}, Pixtransform (\emph{PixT})~\cite{lutio2019guided}, and a version of LGR~\cite{de2022learning} that is based on raw RGB values in the guide, rather than learned features. In much the same way, we also run our diffusion-adjustment scheme with diffusion weights derived from raw RGB values.
For upsampling factors of $\times$4 to $\times$16, the scores are taken directly from \cite{de2022learning}, for scale $\times$32 we have generated all results ourselves, following the setup described in \cite{de2022learning} and using their open-source code base.
As error metrics, we show both the mean squared error (MSE) and the mean absolute error (MAE) of the predicted depth images.

Our experiments were conducted using PyTorch \cite{paszke2019pytorch} and we train all methods, including our own, with the $L_{1}$ loss function. Further details regarding hyper-parameters for training will be provided in the supplementary material and code to ensure reproducibility. For the learning-free variant of DADA we set $\kappa = 0.03$ (see \cref{eq:c}).

\subsection{Results}

\setlength{\tabcolsep}{3pt}
\newlength{\imwc}
\setlength{\imwc}{0.125\textwidth}

\begin{figure}[t]
       \centering
      \footnotesize
    \begin{tabular}{cccc}
    \includegraphics[width=\imwc]{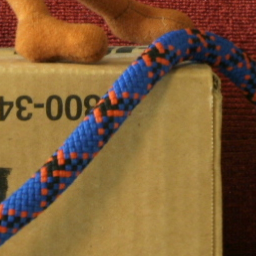} &
    \includegraphics[width=\imwc]{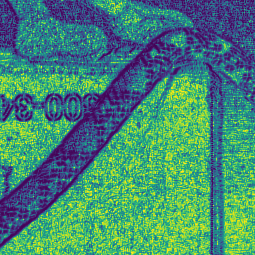} &
    \includegraphics[width=\imwc]{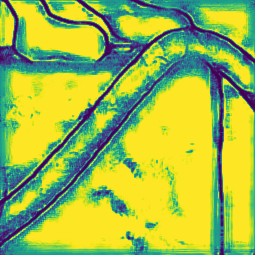} &
    \multirow{2}{*}[1.75cm]{\includegraphics[height=2\imwc]{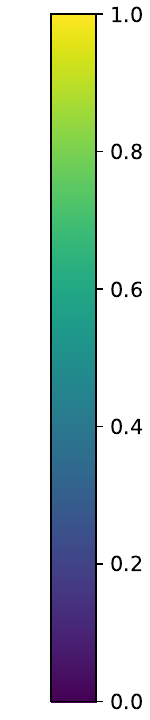} } \\
    \includegraphics[width=\imwc]{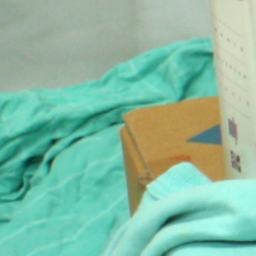} &
    \includegraphics[width=\imwc]{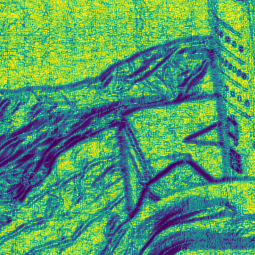} &
    \includegraphics[width=\imwc]{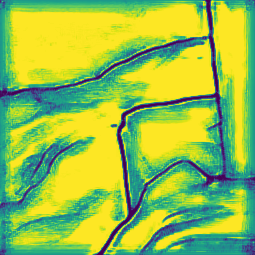} & \\
    Guide & RGB coefficients & Deep coefficients 
    % \vspace{-0.5em}
    % \caption{ Deep Feature }
    \end{tabular}
    \vspace{-0.5em}
    \caption{Diffusion coefficients calculated from raw RGB values and from deep features. The CNN focuses on object discontinuities rather than texture edges. Note the text on the box, top left.}
    \label{fig:diff_coeff}
\end{figure}

\Cref{tab:cmp_l} shows quantitative results on all three datasets for the learned methods, while \cref{tab:cmp_nl} shows the results for the non-learned ones.
Our proposed method obtains the best performance in terms of both MSE and MAE on all three datasets, and across the full range of upsampling factors, with one single exception (at factor $\times$16 on Middlebury, it is 2\textsuperscript{nd}-best).
The results underline the general trend towards learned, high-level feature representations: generally speaking, the learned methods clearly outperform the learning-free ones. In particular, for LGR as well as for our DADA, the versions with learned features easily beat their RGB-based counterparts.
Having said that, we note that many learning-based methods seem to struggle with large-scale differences between the source and the guide (respectively, the target): the gap between learned and non-learned methods shrinks with increasing upsampling factors, and in fact, the performance of several dedicated super-resolution methods (learning-free ones, but also learned ones) falls below that of na\"ive bicubic upsampling. 

\begin{figure}[th]
  \centering
  \begin{subfigure}{0.82\linewidth}
    \includegraphics[width=\textwidth]{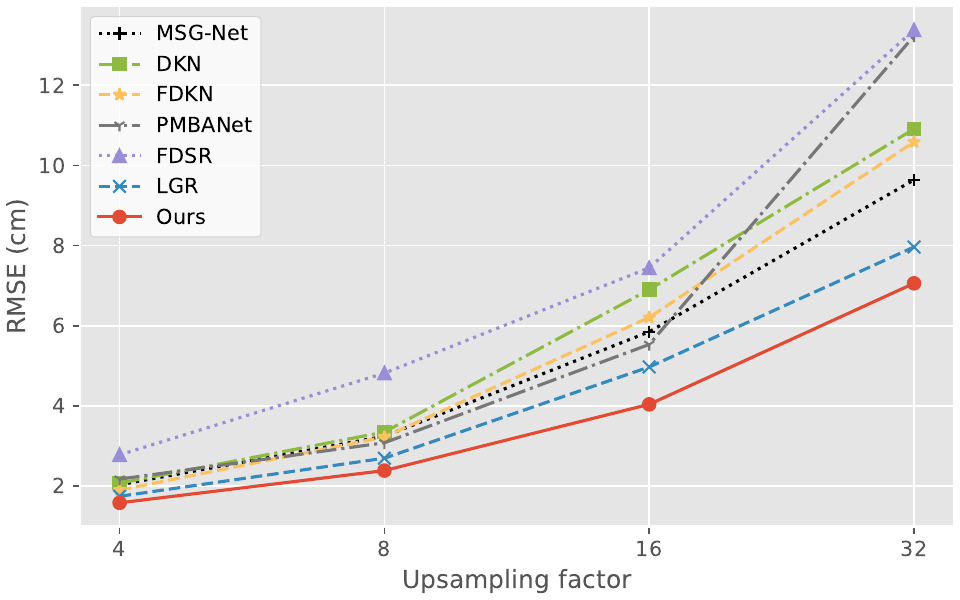}
    \caption{RMSE (cm) of learning-based methods}
    \label{fig:curves_middlebury_sup}
  \end{subfigure}
  
  \vspace{6pt}
  
  \begin{subfigure}{0.82\linewidth}
    \includegraphics[width=\textwidth]{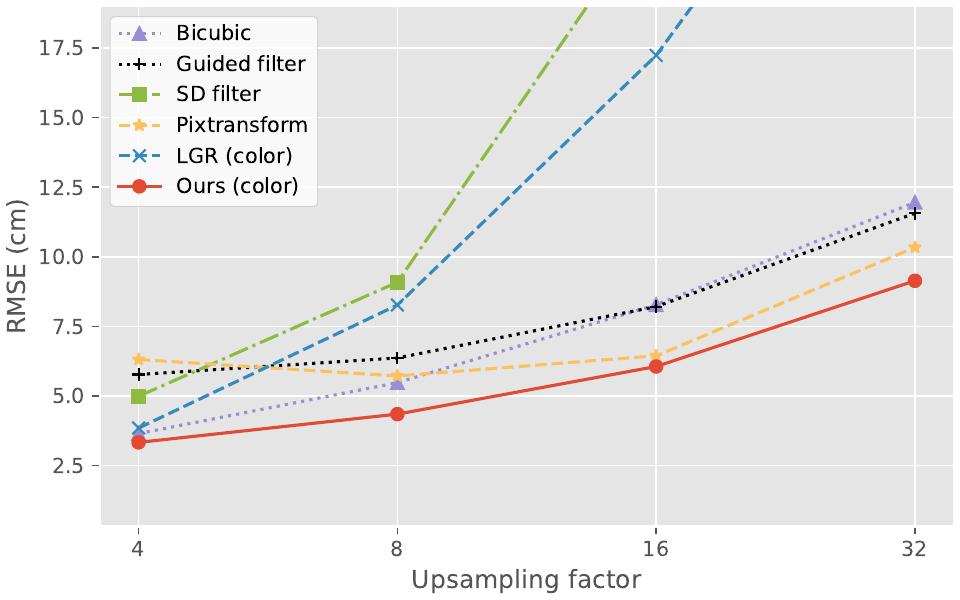}
    \caption{RMSE (cm) of learning-free methods}
    \label{fig:curves_middlebury_unsup}
  \end{subfigure}
  \caption{RMSE values at different upsampling factors, for the Middlebury dataset. Our method consistently achieves the lowest error.}
  \label{fig:curves_middlebury}
\end{figure}

RMSE curves for different scaling factors using the Middlebury dataset can be seen in \cref{fig:curves_middlebury}. The plots for the other datasets follow the same trends and will be included in the supplementary material. As seen in these results, the performances of several of the considered methods degrade severely for larger scaling factors. In fact, for scale $\times$32 many of the methods perform worse than simple bicubic interpolation. The results also show that for the lowest scaling factor ($\times$4) many learned methods achieve similar performance, suggesting that guided super-resolution for moderate scale factors is reaching saturation, at least \wrt the currently available training and test data.

\Cref{fig:depth_upsampling} contains results from several of the compared methods on all three datasets with scaling factors of $\times$16 and $\times$32. The residual images make it clear that DADA is superior to all other methods, including LGR -- the previous best. Strict constraint to the source is especially advantageous in regions with no depth discontinuities. DADA is nevertheless able to produce sharp edges where necessary.
Encircled depth discontinuities seem to be the most challenging cases for all methods. Although DADA still outperforms other methods, those situations still offer room for improvement.

\begin{table}[t]
    \centering
    \footnotesize
    % \resizebox{0.475\textwidth}{!}{%
    \begin{tabularx}{0.45\textwidth}{ll | YYY | YYY}
     &  & \multicolumn{3}{c|}{\textbf{DIML}} & \multicolumn{3}{c}{\textbf{Middlebury}} \\
     &  & MSE & MAE & MSE$^\ddag$ & MSE & MAE & MSE$^\ddag$ \\ \midrule \midrule
    \parbox[t]{2mm}{\multirow{6}{*}{\rotatebox[origin=c]{90}{$\times$8}}}\hspace{5pt} & MSG  & 5.76 & 0.51 & 6.16  & 11.0 & 0.54 & 5.01 \\
     & FDKN          & 6.74 & 0.53 & 0.20  & 10.0 & 0.43 & 0.20 \\
     & PMBA          & 7.35 & 0.59 & 0.04  & 9.62 & 0.46 & 0.06 \\ 
     & FDSR          & 7.73 & 0.74 & 0.45  & 18.4 & 0.73 & 7.20 \\
     & LGR           & 4.95 & 0.40 & 0.002 & 8.25 & \textbf{0.35} & 0.001 \\
     & \textbf{DADA}\hspace{5pt} & \textbf{4.58} & \textbf{0.38} & \textbf{0.0}   & \textbf{7.82} & \textbf{0.35} & \textbf{0.0} \\ \midrule
    \parbox[t]{2mm}{\multirow{6}{*}{\rotatebox[origin=c]{90}{$\times$32}}}\hspace{5pt} & MSG & 60.2 & 2.82 & 0.59 & 68.5 & 2.31 & 0.46 \\
     & FDKN          & 97.2 & 3.91 & 1.09 & 98.3 & 3.05 & 0.68 \\
     & PMBA          & 312 & 4.96 & 1.24 & 118 & 3.71 & 6.62 \\ 
     & FDSR          & 239 & 6.45 & 3.70 & 177 & 4.67 & 2.47 \\
     & LGR           & 64.5 & 2.83 & 0.001 & 70.7 & 2.28 & 0.008 \\
     & \textbf{DADA} & \textbf{56.9} & \textbf{2.35} & \textbf{0.0} & \textbf{52.3} & \textbf{1.67} & \textbf{0.0} \\

    \end{tabularx}%
    % }%\usepackage{adjustbox}
    \caption{Cross-dataset generalization performance. All learned methods were trained on NYUv2, then tested on DIML and Middlebury, with scaling factors of $\times$8 and $\times$32. Errors are in $\text{cm}^ 2$ for the MSE and in cm for the MAE. The low-resolution MSE$^\ddag$, \ie, after downsampling the predicted target, indicates inconsistency with the forward model (linear downsampling).
    }
    \vspace{-0.3cm}
\label{tab:cmp_crosstesting}
\end{table}

Our quantitative and qualitative results both show the superiority of the results DADA yields when compared to all other methods. Our results also lend further support to the findings of LGR~\cite{de2022learning} that theoretically founded, optimization-based methods still offer excellent performance within-domain and especially out-of-domain if (and only if) they are integrated with the large receptive field and superior perceptual grouping abilities of modern, deep feature extractors. 
Having said that, we find that DADA surpasses LGR in terms of both reconstruction error and computational cost (memory consumption as well as runtime), especially at high upsampling factors.

In \Cref{fig:diff_coeff} we show a pair of diffusion coefficients derived from deep features and contrast them with coefficients that are directly derived from the RGB values. The learned coefficients are more robust to printed text and logos, image noise, and textured surfaces, whereas the unlearned ones are overly sensitive to those.

\newcommand\MID{6}
\newcommand\MIDbis{400}

\newcommand\NYUv{3}
\newcommand\NYUvbis{97}

\newcommand\DIMLv{200}
\newcommand\DIMLvbis{1}

\setlength{\tabcolsep}{1pt}
\newlength{\imw}
\setlength{\imw}{0.083\textwidth}

\begin{figure*}[t]
\vspace{-0.2cm}
  \centering
  \footnotesize
\begin{tabular}{ccccccccccccc}
\multirow{2}{*}[0.35cm]{\rotatebox{90}{\bf Middlebury}} &
% \parbox[t]{2mm}{\multirow{2}{*}{\rotatebox[origin=c]{90}{\textbf{Middlebury}}}} &
\rotatebox[origin=c]{90}{$\times$16} &
\includegraphics[width=\imw]{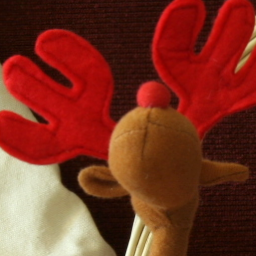} &
\includegraphics[width=\imw]{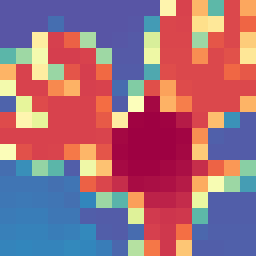} &
\includegraphics[width=\imw]{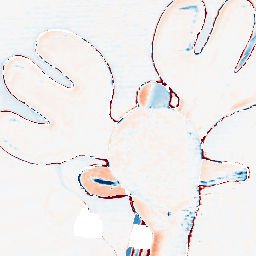} &
\includegraphics[width=\imw]{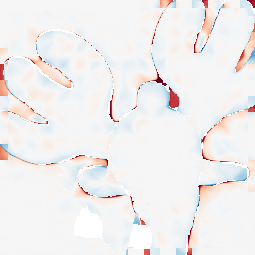} &
\includegraphics[width=\imw]{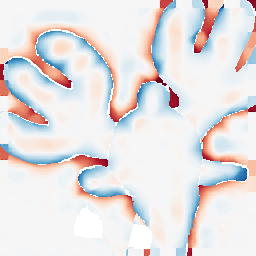} &
\includegraphics[width=\imw]{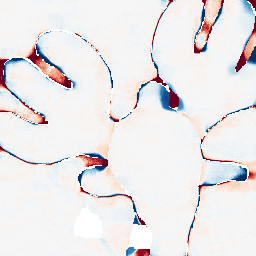} &
\includegraphics[width=\imw]{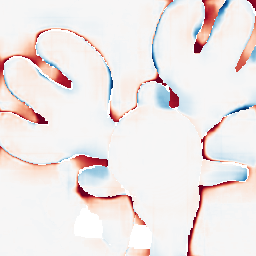} &
\includegraphics[width=\imw]{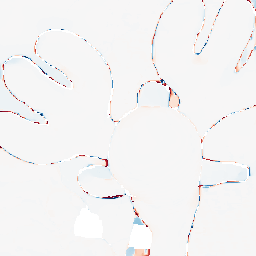} &
\includegraphics[width=\imw]{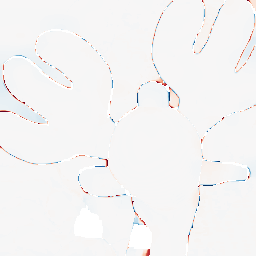} &
\includegraphics[width=\imw]{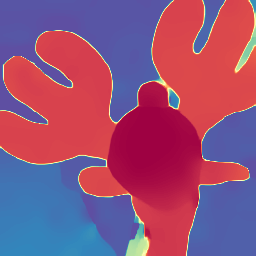} &
\includegraphics[width=\imw]{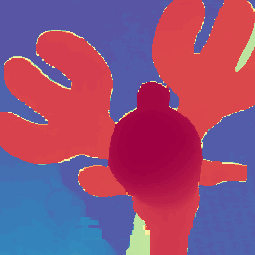}\\
& \rotatebox{90}{$\times$32} &
\includegraphics[width=\imw]{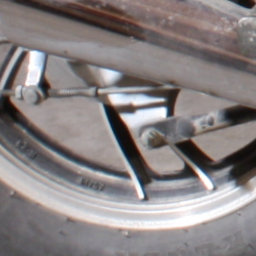} &
\includegraphics[width=\imw]{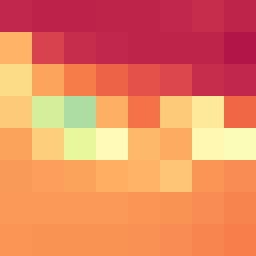} &
\includegraphics[width=\imw]{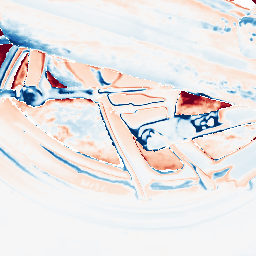} &
\includegraphics[width=\imw]{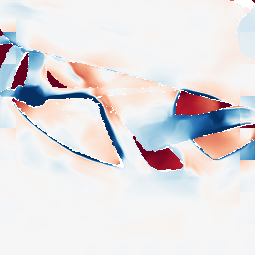} &
\includegraphics[width=\imw]{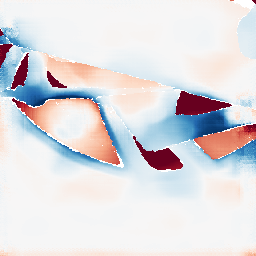} &
\includegraphics[width=\imw]{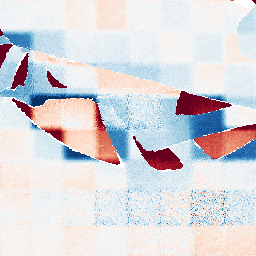} &
\includegraphics[width=\imw]{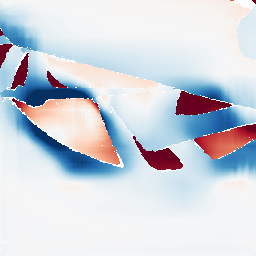} &
\includegraphics[width=\imw]{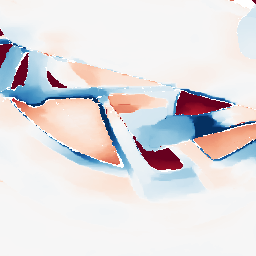} &
\includegraphics[width=\imw]{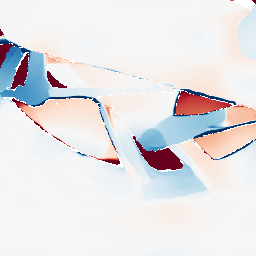} &
\includegraphics[width=\imw]{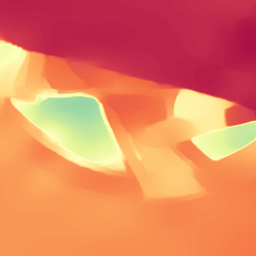} &
\includegraphics[width=\imw]{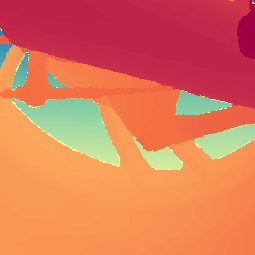} \\
\midrule
\multirow{2}{*}{\rotatebox{90}{\bf NYUv2}} & \rotatebox{90}{$\times$16} &
\includegraphics[width=\imw]{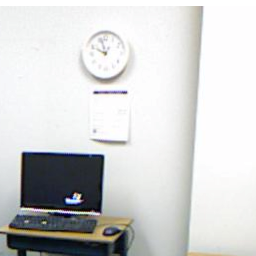} &
\includegraphics[width=\imw]{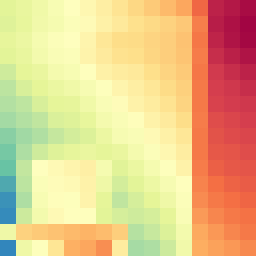} &
\includegraphics[width=\imw]{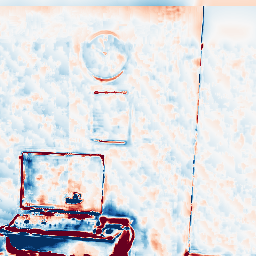} &
\includegraphics[width=\imw]{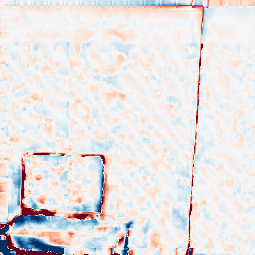} &
\includegraphics[width=\imw]{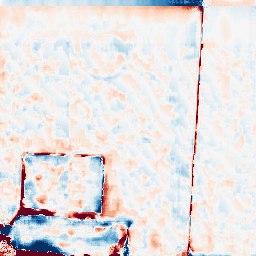} &
\includegraphics[width=\imw]{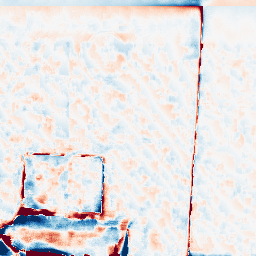} &
\includegraphics[width=\imw]{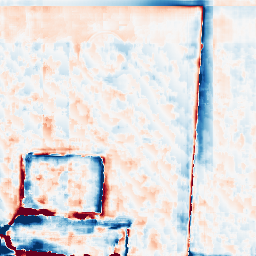} &
\includegraphics[width=\imw]{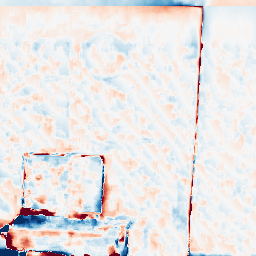} &
\includegraphics[width=\imw]{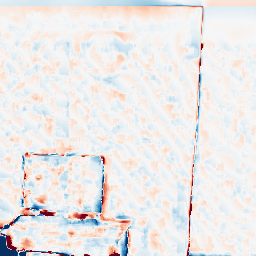} &
\includegraphics[width=\imw]{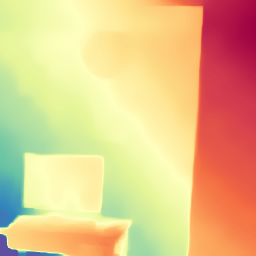} &
\includegraphics[width=\imw]{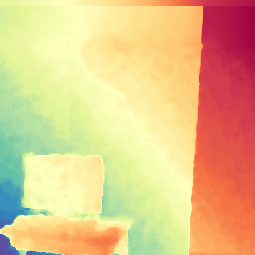}\\
& \rotatebox{90}{$\times$32} &
\includegraphics[width=\imw]{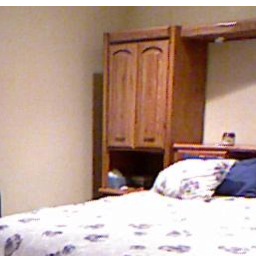} &
\includegraphics[width=\imw]{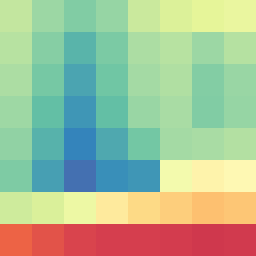} &
\includegraphics[width=\imw]{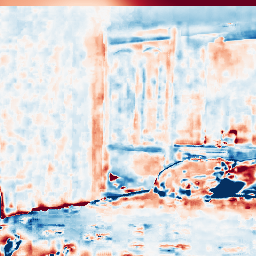} &
\includegraphics[width=\imw]{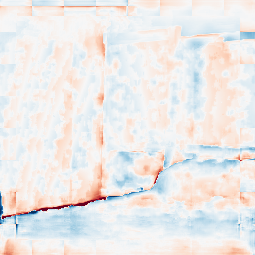} &
\includegraphics[width=\imw]{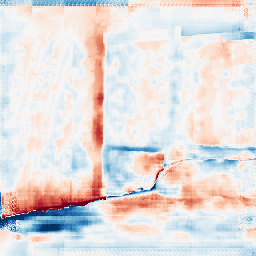} &
\includegraphics[width=\imw]{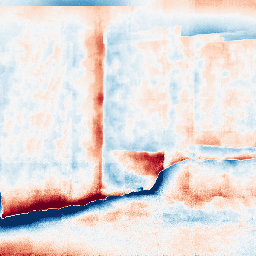} &
\includegraphics[width=\imw]{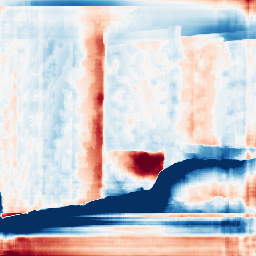} &
\includegraphics[width=\imw]{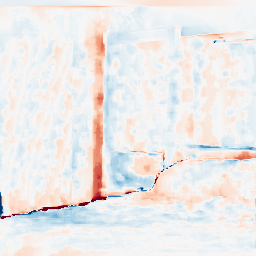} &
\includegraphics[width=\imw]{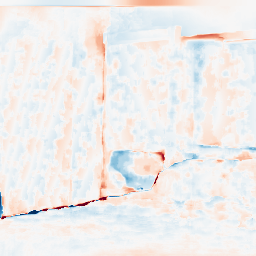} &
\includegraphics[width=\imw]{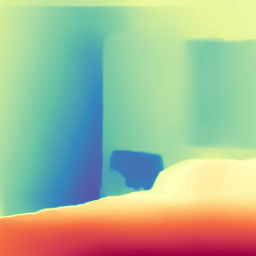} &
\includegraphics[width=\imw]{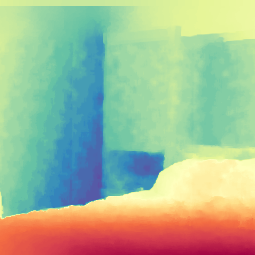}\\
\midrule
\multirow{2}{*}{\rotatebox{90}{\bf DIML}} & \rotatebox{90}{$\times$16} &
\includegraphics[width=\imw]{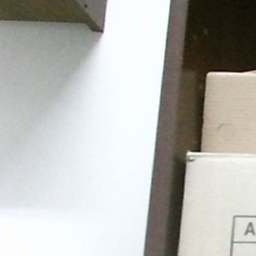} &
\includegraphics[width=\imw]{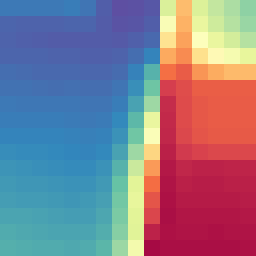} &
\includegraphics[width=\imw]{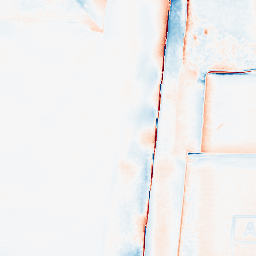} &
\includegraphics[width=\imw]{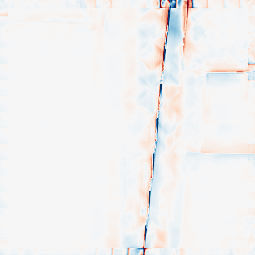} &
\includegraphics[width=\imw]{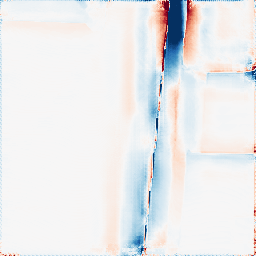} &
\includegraphics[width=\imw]{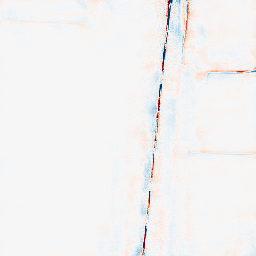} &
\includegraphics[width=\imw]{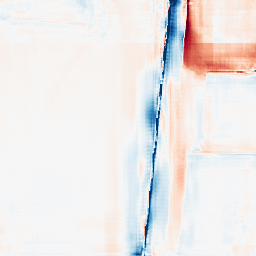} &
\includegraphics[width=\imw]{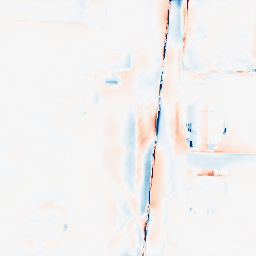} &
\includegraphics[width=\imw]{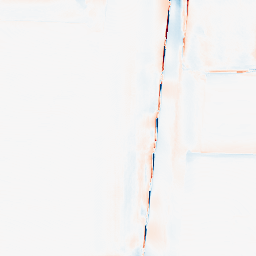} &
\includegraphics[width=\imw]{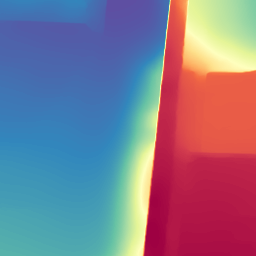} &
\includegraphics[width=\imw]{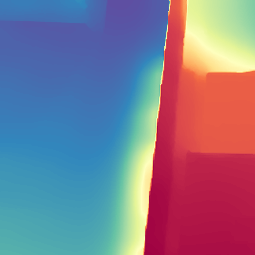}\\
& \rotatebox{90}{$\times$32} &
\includegraphics[width=\imw]{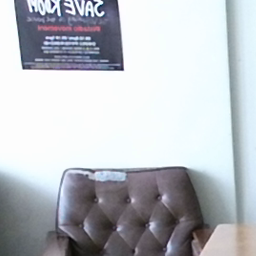} &
\includegraphics[width=\imw]{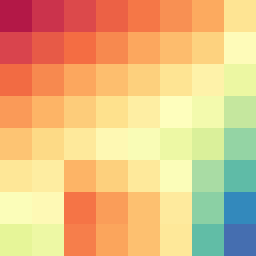} &
\includegraphics[width=\imw]{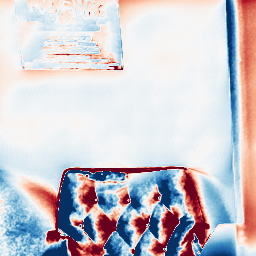} &
\includegraphics[width=\imw]{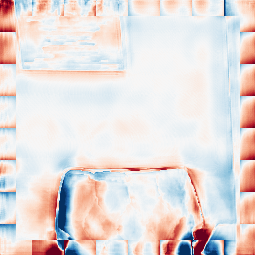} &
\includegraphics[width=\imw]{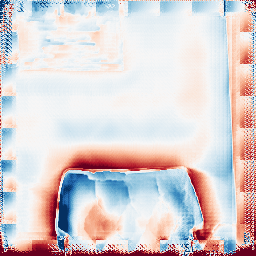} &
\includegraphics[width=\imw]{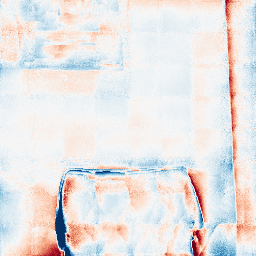} &
\includegraphics[width=\imw]{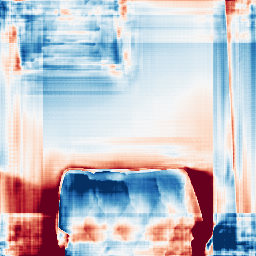} &
\includegraphics[width=\imw]{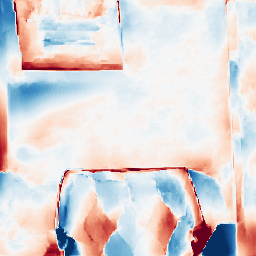} &
\includegraphics[width=\imw]{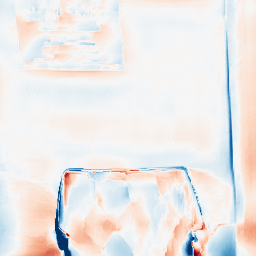} &
\includegraphics[width=\imw]{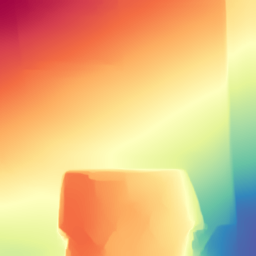} &
\includegraphics[width=\imw]{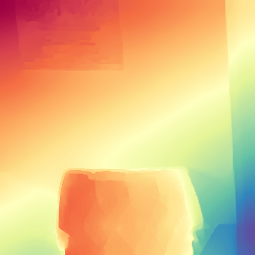}\\
& & Guide & Source & PixT & MSG & FDKN & PMBA & FDSR & LGR & DADA & DADA pred. & GT
\end{tabular}
\vspace{-0.5em}
\caption{Error maps for different guided super-resolution methods. Blue denotes under-estimated depth, red denotes over-estimation. All errors maps in a row have the same color scale. The last two columns juxtapose our predictions and the ground truth. }
\label{fig:depth_upsampling}
\end{figure*}

Experience suggests that deep learning methods tend to (over-)fit to dataset-specific characteristics and generalize poorly from one dataset to another.
For LGR, which like our method explicitly enforces consistency between the (downsampled) target and the source, the authors in fact demonstrate improved generalization compared to purely learned frameworks. They attribute this behavior to the consistency constraint, which is data-independent and therefore not affected by domain gaps \cite{de2022learning}.
We have also explored the generalization of our method across datasets, and report the results in \Cref{tab:cmp_crosstesting}. 
For these experiments, we train on the NYUv2 dataset for a super-resolution factor of $\times$8 and $\times$32, and test the resulting network on Middlebury and DIML. Also for this task, our method outperforms all competitors, including LGR, which ranks second.

\begin{figure*}[ht]
    \centering
    
    \begin{subfigure}{0.32\linewidth}
        \centering
        \hspace*{-20pt}\includegraphics[width=\textwidth]{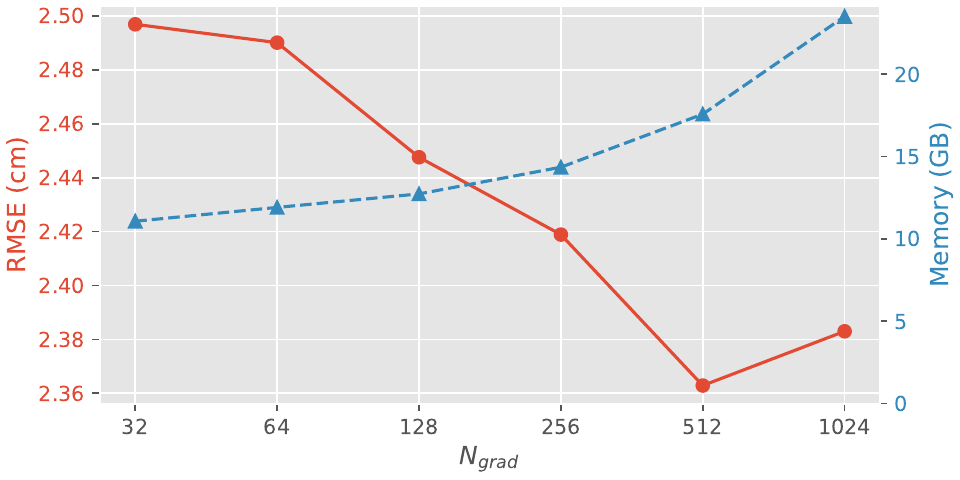}
        \caption{Number $N_\text{grad}$ of iterations with gradient propagation and memory consumption.}
        \label{fig:ablation_Ntrain}
    \end{subfigure}
    \hfill
    \begin{subfigure}{0.32\linewidth}
        \centering
        \hspace*{-20pt}\includegraphics[width=\textwidth]{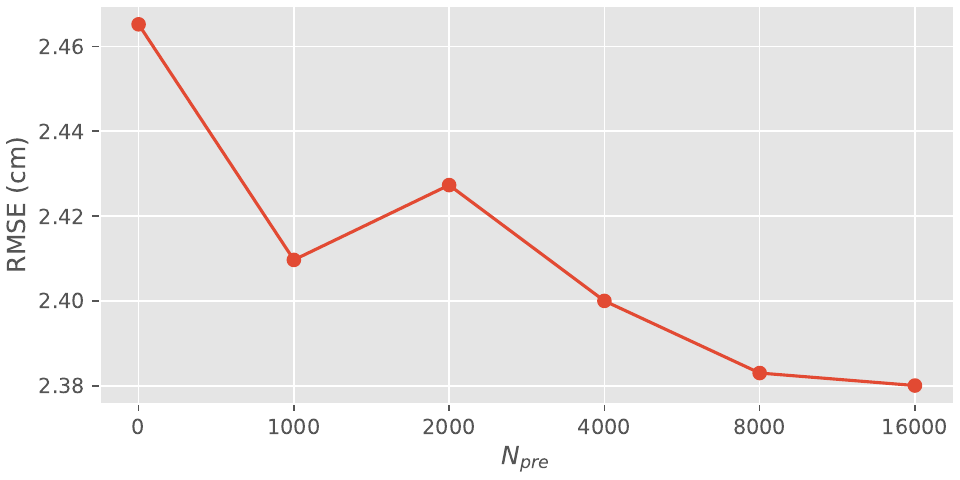}
        \caption{Number $N_\text{pre}$ of iterations without gradient propagation.}
        \label{fig:ablation_Npre}
    \end{subfigure}
    \hfill
    \begin{subfigure}{0.32\linewidth}
        \centering
        \includegraphics[width=\textwidth]{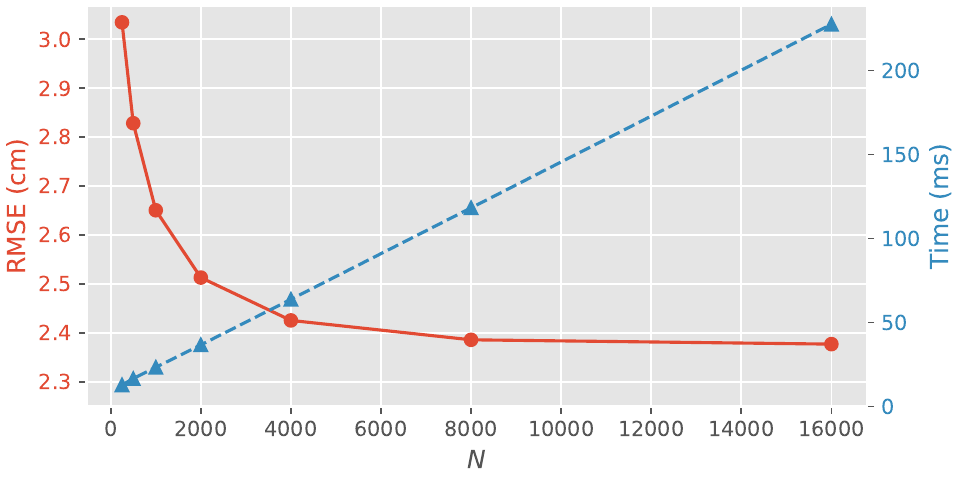}
        \caption{Number of iterations used at inference time and average computing time per sample.}
        \label{fig:ablation_Ninference}
    \end{subfigure}
    \caption{Plots for ablation experiments that explore the effects of different numbers of iterations in the diffusion-adjustment loop. We generally see an increase in performance for larger numbers of iterations, albeit at higher computational cost.}
    \label{fig:ablations}
\end{figure*}

\subsection{Ablations}

In this section, we experimentally study the effects of different parameters and modules on the results of our super-resolution method. We run these experiments on the Middlebury dataset with upsampling factor $\times$8, which we found to be a representative setting. The memory requirements reported below always refer to a training batch consisting of 8 images of size $256 \times 256$.

\paragraph{Initialization.} Our default initialization scheme for the super-resolved image (bicubic upsampling of the guide) and for the feature extractor (Imagenet pretraining) are not the only options. Compared to the base case ($MSE=5.67$, $MAE=0.199$), we explore the two simplifications: (1) random initialization of the ResNet-50 backbone weights instead of ImageNet pretraining (which yields ($MSE=5.92$, $MAE=0.211$); and (2) constant initialization of $\mathbf{Y}_0$ rather than bicubic interpolation of $\mathbf{S}$ (which yields ($MSE=5.64$, $MAE=0.200$). In both cases, we observe only tiny differences \wrt the base case. With regard to the backbone weights, this suggests that the  available amount of training data is sufficient to learn a suitable feature representation. For the initialization of the target image $\mathbf{Y}_0$, it is not surprising that this has almost no effect on the result. It is easy to see that the first diffusion step will not alter the constant input and that the output $\mathbf{Y}_1$ of the first adjustment step will then be the nearest-neighbor upsampling of $\mathbf{S}$. Meaning that after a single iteration of the diffusion-adjustment loop, the constant initialization has largely caught up with the seemingly more sophisticated bicubic one (and in fronto-parallel areas perhaps even has overtaken it).

\paragraph{Number of training iterations with gradient.} As explained above, due to memory constraints we limit the number of diffusion iterations that propagate gradients into the feature extractor.
In \Cref{fig:ablation_Ntrain}, we explore how this affects feature learning. Memory requirements increase with $N_{\text{grad}}$, ranging from 11 GB for $N_{\text{grad}} = 32$ up to 23 GB for $N_{\text{grad}} = 1024$. We find that larger values of $N_{\text{grad}}$ generally lead to better performance, likely due to the gradients being more representative of the equilibrium point of the diffusion-adjustment process.

\paragraph{Number of training iterations without gradient.} A main hyper-parameter of the (asymptotic) diffusion process is the total number of iterations. For a fixed setting of $N_\text{grad}$ this is, in our scheme, governed by the number of preceding gradient-free iterations, $N_\text{pre}$. We vary that value and once again we find that, as expected, larger values generally lead to better results, likely because the gradients are computed from a state closer to the equilibrium point. We see little further gain after $N_{\text{pre}} = 8000$.

\paragraph{Number of iterations at test time.} Once trained, we are free to decide for how long we run the diffusion process at inference time. In \Cref{fig:ablation_Ninference} we explore how the number of iterations at test time impacts the results. Once more, higher iteration counts $N$ lead to better results, as the process converges further towards the equilibrium point $\lim_{t \to \infty} \mathbf{Y}_t$. We see little improvement for $N>8000$. As computing time (also shown in the graph) scales linearly with $N$, we see no reason to go beyond $N=8000$.

\section{Conclusion}

We have presented DADA, a simple yet extremely effective approach to guided super-resolution of depth images. The method links guided anisotropic diffusion with deep learning. The diffusion part offers a computationally efficient optimization framework to optimally align the prediction with the individual input images at test time and explicitly constrains the results to comply with the source image. The deep learning-based feature extractor harnesses the representation power of CNNs to optimally inform the diffusion process, by injecting context cues collected over large receptive fields into the diffusion weights.
In our experiments, DADA reaches top performance on three popular datasets and outperforms the prior state-of-the-art, including learning-based, optimization-based as well as hybrid methods.
We also found that DADA is fairly robust in terms of engineering details: it is barely affected by the choice of pretraining and initialization scheme and is relatively tolerant against hyper-parameter changes.

An interesting direction for future work will be to formulate the diffusion-adjustment iteration as an ordinary differential equation.
%so that it can be solved with the adjoint method~\cite{pontryagin1987mathematical}.
This could potentially make it possible to back-propagate through the entire diffusion process with the adjoint sensitivity method~\cite{pontryagin1987mathematical}, rather than unrolling a limited number of iterations. 
At a conceptual level, we hope that our work motivates further research about the combination of classical optimization-based computer vision and modern, deep learning-based image representations. In line with others~\cite{riegler2016deep, de2022learning}, our work suggests that such hybrid methods hold great potential, not only for super-resolution, but also for other low-level vision tasks, and perhaps beyond.

%\clearpage
%%%%%%%%% REFERENCES
{\small
\bibliographystyle{ieee_fullname}
\bibliography{refs}

\begin{thebibliography}{10}\itemsep=-1pt

\bibitem{almasri2018rgb}
Feras Almasri and Olivier Debeir.
\newblock {RGB} guided thermal super-resolution enhancement.
\newblock {\em Int'l Conf on Cloud Computing Technologies and Applications},
  2018.

\bibitem{daudt2019guided}
Rodrigo Caye~Daudt, Bertrand Le~Saux, Alexandre Boulch, and Yann Gousseau.
\newblock Guided anisotropic diffusion and iterative learning for weakly
  supervised change detection.
\newblock {\em CVPR Workshops}, 2019.

\bibitem{DIML4}
Jaehoon Cho, Dongbo Min, Youngjung Kim, and Kwanghoon Sohn.
\newblock Deep monocular depth estimation leveraging a large-scale outdoor
  stereo dataset.
\newblock {\em Expert Systems with Applications}, 178:114877, 2021.

\bibitem{daudt2021weakly}
Rodrigo~Caye Daudt, Bertrand Le~Saux, Alexandre Boulch, and Yann Gousseau.
\newblock Weakly supervised change detection using guided anisotropic
  diffusion.
\newblock {\em Machine Learning}, pages 1--27, 2021.

\bibitem{de2022learning}
Riccardo {de Lutio}, Alexander Becker, Stefano {D'Aronco}, Stefania Russo,
  Jan~D Wegner, and Konrad Schindler.
\newblock Learning graph regularisation for guided super-resolution.
\newblock {\em CVPR}, 2022.

\bibitem{lutio2019guided}
Riccardo {de Lutio}, Stefano {D'Aronco}, Jan~Dirk Wegner, and Konrad Schindler.
\newblock Guided super-resolution as pixel-to-pixel transformation.
\newblock {\em ICCV}, 2019.

\bibitem{deng2009imagenet}
Jia Deng, Wei Dong, Richard Socher, Li-Jia Li, Kai Li, and Li Fei-Fei.
\newblock Imagenet: A large-scale hierarchical image database.
\newblock {\em CVPR}, 2009.

\bibitem{diebel2005application}
James Diebel and Sebastian Thrun.
\newblock An application of {Markov} random fields to range sensing.
\newblock {\em NIPS}, 2005.

\bibitem{eichhardt2017image}
Ivan Eichhardt, Dmitry Chetverikov, and Zsolt Janko.
\newblock Image-guided {ToF} depth upsampling: a survey.
\newblock {\em Machine Vision and Applications}, 28(3):267--282, 2017.

\bibitem{ham2017robust}
Bumsub Ham, Minsu Cho, and Jean Ponce.
\newblock Robust guided image filtering using nonconvex potentials.
\newblock {\em IEEE TPAMI}, 40(1):192--207, 2017.

\bibitem{he2012guided}
Kaiming He, Jian Sun, and Xiaoou Tang.
\newblock Guided image filtering.
\newblock {\em IEEE TPAMI}, 35(6):1397--1409, 2012.

\bibitem{he2016deep}
Kaiming He, Xiangyu Zhang, Shaoqing Ren, and Jian Sun.
\newblock Deep residual learning for image recognition.
\newblock {\em CVPR}, 2016.

\bibitem{he2021}
Lingzhi He, Hongguang Zhu, Feng Li, Huihui Bai, Runmin Cong, Chunjie Zhang,
  Chunyu Lin, Meiqin Liu, and Yao Zhao.
\newblock Towards fast and accurate real-world depth super-resolution:
  Benchmark dataset and baseline.
\newblock {\em CVPR}, 2021.

\bibitem{Hirschmuller2007}
Heiko Hirschm{\"u}ller and Daniel Scharstein.
\newblock Evaluation of cost functions for stereo matching.
\newblock {\em CVPR}, 2007.

\bibitem{Tak-Wai2016}
Tak-Wai Hui, Chen~Change Loy, and Xiaoou Tang.
\newblock Depth map super-resolution by deep multi-scale guidance.
\newblock {\em ECCV}, 2016.

\bibitem{izraelevitz1994}
David Izraelevitz.
\newblock Model-based multispectral sharpening.
\newblock {\em Proceedings of SPIE}, 1994.

\bibitem{keys1981cubic}
Robert Keys.
\newblock Cubic convolution interpolation for digital image processing.
\newblock {\em IEEE T Acoustics, Speech, and Signal processing},
  29(6):1153--1160, 1981.

\bibitem{Kim2021}
Beomjun Kim, Jean Ponce, and Bumsub Ham.
\newblock Deformable kernel networks for joint image filtering.
\newblock {\em IJCV}, 129(2):579--600, 2021.

\bibitem{DIML2}
Sunok Kim, Dongbo Min, Bumsub Ham, Seungryong Kim, and Kwanghoon Sohn.
\newblock Deep stereo confidence prediction for depth estimation.
\newblock {\em ICIP}, 2017.

\bibitem{DIML1}
Youngjung Kim, Bumsub Ham, Changjae Oh, and Kwanghoon Sohn.
\newblock Structure selective depth superresolution for {RGB-D} cameras.
\newblock {\em IEEE TIP}, 25(11):5227--5238, 2016.

\bibitem{DIML3}
Youngjung Kim, Hyungjoo Jung, Dongbo Min, and Kwanghoon Sohn.
\newblock Deep monocular depth estimation via integration of global and local
  predictions.
\newblock {\em IEEE TIP}, 27(8):4131--4144, 2018.

\bibitem{kingma2014adam}
Diederik~P Kingma and Jimmy Ba.
\newblock Adam: A method for stochastic optimization.
\newblock {\em arXiv preprint arXiv:1412.6980}, 2014.

\bibitem{lanaras2018super}
Charis Lanaras, Jos{\'e} Bioucas-Dias, Silvano Galliani, Emmanuel Baltsavias,
  and Konrad Schindler.
\newblock Super-resolution of {Sentinel-2} images: Learning a globally
  applicable deep neural network.
\newblock {\em ISPRS J Photogrammetry and Remote Sensing}, 146:305--319, 2018.

\bibitem{liu2013guided}
Junyi Liu and Xiaojin Gong.
\newblock Guided depth enhancement via anisotropic diffusion.
\newblock {\em Pacific-Rim Conf on Multimedia}, 2013.

\bibitem{ma2013constant}
Ziyang Ma, Kaiming He, Yichen Wei, Jian Sun, and Enhua Wu.
\newblock Constant time weighted median filtering for stereo matching and
  beyond.
\newblock {\em ICCV}, 2013.

\bibitem{mac2012patch}
Oisin {Mac Aodha}, Neill D.~F. Campbell, Arun Nair, and Gabriel~J. Brostow.
\newblock Patch based synthesis for single depth image super-resolution.
\newblock {\em ECCV}, 2012.

\bibitem{min2011depth}
Dongbo Min, Jiangbo Lu, and Minh~N Do.
\newblock Depth video enhancement based on weighted mode filtering.
\newblock {\em IEEE TIP}, 21(3):1176--1190, 2011.

\bibitem{park2019planar}
Min-Gyu Park and Kuk-Jin Yoon.
\newblock As-planar-as-possible depth map estimation.
\newblock {\em Computer Vision and Image Understanding}, 181:50--59, 2019.

\bibitem{paszke2019pytorch}
Adam Paszke, Sam Gross, Francisco Massa, Adam Lerer, James Bradbury, Gregory
  Chanan, Trevor Killeen, Zeming Lin, Natalia Gimelshein, Luca Antiga, et~al.
\newblock {PyTorch}: An imperative style, high-performance deep learning
  library.
\newblock {\em NeurIPS}, 2019.

\bibitem{patterson1992}
Tim~J. Patterson, Michael~E. Bullock, and Alan~D. Wada.
\newblock Multispectral band sharpening using pseudoinverse estimation and
  fuzzy reasoning.
\newblock {\em Proceedings of SPIE}, 1992.

\bibitem{perona1990scale}
Pietro Perona and Jitendra Malik.
\newblock Scale-space and edge detection using anisotropic diffusion.
\newblock {\em IEEE TPAMI}, 12(7):629--639, 1990.

\bibitem{pontryagin1987mathematical}
Lev~S. Pontryagin.
\newblock {\em Mathematical theory of optimal processes}.
\newblock CRC Press, 1987.

\bibitem{riegler2016deep}
Gernot Riegler, David Ferstl, Matthias R{\"u}ther, and Horst Bischof.
\newblock A deep primal-dual network for guided depth super-resolution.
\newblock {\em BMVC}, 2016.

\bibitem{ronneberger2015u}
Olaf Ronneberger, Philipp Fischer, and Thomas Brox.
\newblock U-net: Convolutional networks for biomedical image segmentation.
\newblock {\em MICCAI}, 2015.

\bibitem{rossi2020joint}
Mattia Rossi, Mireille~El Gheche, Andreas Kuhn, and Pascal Frossard.
\newblock Joint graph-based depth refinement and normal estimation.
\newblock In {\em Proceedings of the IEEE/CVF Conference on Computer Vision and
  Pattern Recognition}, pages 12154--12163, 2020.

\bibitem{Scharstein2014}
Daniel Scharstein, Heiko Hirschm{\"u}ller, York Kitajima, Greg Krathwohl, Nera
  Ne{\v{s}}i{\'{c}}, Xi Wang, and Porter Westling.
\newblock High-resolution stereo datasets with subpixel-accurate ground truth.
\newblock {\em GCPR}, 2014.

\bibitem{Scharstein2007}
Daniel Scharstein and Chris Pal.
\newblock Learning conditional random fields for stereo.
\newblock {\em CVPR}, 2007.

\bibitem{Scharstein2003}
Daniel Scharstein and Richard Szeliski.
\newblock High-accuracy stereo depth maps using structured light.
\newblock {\em CVPR}, 2001.

\bibitem{Scharstein2001}
Daniel Scharstein, Richard Szeliski, and Ramin Zabih.
\newblock A taxonomy and evaluation of dense two-frame stereo correspondence
  algorithms.
\newblock {\em IJCV}, 47(1):7--42, 2002.

\bibitem{Shelhamer2017}
Evan Shelhamer, Jonathan Long, and Trevor Darrell.
\newblock Fully convolutional networks for semantic segmentation.
\newblock {\em IEEE TPAMI}, 2017.

\bibitem{song2020channel}
Xibin Song, Yuchao Dai, Dingfu Zhou, Liu Liu, Wei Li, Hongdong Li, and Ruigang
  Yang.
\newblock Channel attention based iterative residual learning for depth map
  super-resolution.
\newblock {\em CVPR}, 2020.

\bibitem{uezato2020guided}
Tatsumi Uezato, Danfeng Hong, Naoto Yokoya, and Wei He.
\newblock Guided deep decoder: Unsupervised image pair fusion.
\newblock {\em ECCV}, 2020.

\bibitem{ulyanov2018deep}
Dmitry Ulyanov, Andrea Vedaldi, and Victor Lempitsky.
\newblock Deep image prior.
\newblock {\em CVPR}, 2018.

\bibitem{wang2019multi}
Jin Wang, Wei Xu, Jian-Feng Cai, Qing Zhu, Yunhui Shi, and Baocai Yin.
\newblock Multi-direction dictionary learning based depth map super-resolution
  with autoregressive modeling.
\newblock {\em IEEE T Multimedia}, 22(6):1470--1484, 2019.

\bibitem{wen2018deep}
Yang Wen, Bin Sheng, Ping Li, Weiyao Lin, and David~Dagan Feng.
\newblock Deep color guided coarse-to-fine convolutional network cascade for
  depth image super-resolution.
\newblock {\em IEEE TIP}, 28(2):994--1006, 2018.

\bibitem{xie2015edge}
Jun Xie, Rogerio~Schmidt Feris, and Ming-Ting Sun.
\newblock Edge-guided single depth image super resolution.
\newblock {\em IEEE TIP}, 25(1):428--438, 2015.

\bibitem{yang2007spatial}
Qingxiong Yang, Ruigang Yang, James Davis, and David Nist{\'e}r.
\newblock Spatial-depth super resolution for range images.
\newblock {\em CVPR}, 2007.

\bibitem{Ye2020}
Xinchen Ye, Baoli Sun, Zhihui Wang, Jingyu Yang, Rui Xu, Haojie Li, and Baopu
  Li.
\newblock {PMBANet}: Progressive multi-branch aggregation network for scene
  depth super-resolution.
\newblock {\em IEEE TIP}, 2020.

\bibitem{zhang2018longitudinally}
Yongqin Zhang, Feng Shi, Jian Cheng, Li Wang, Pew-Thian Yap, and Dinggang Shen.
\newblock Longitudinally guided super-resolution of neonatal brain magnetic
  resonance images.
\newblock {\em IEEE T Cybernetics}, 49(2):662--674, 2018.

\end{thebibliography}
}

% \FloatBarrier
% \newpage
\clearpage
% \input{supplementary.tex}

%%%%%%%%% TITLE - 

\twocolumn[\Large \centering \textbf{Supplementary Material} \\\textbf{Guided Depth Super-Resolution by Deep Anisotropic Diffusion} \vspace{1cm}]

\appendix

\renewcommand{\thefigure}{A\arabic{figure}}
\renewcommand{\thetable}{A\arabic{table}}
\setcounter{figure}{0}
\setcounter{table}{0}

\section{Videos}

Videos depicting the diffusing process ($Y_t$ and residuals) are included with the supplementary materials. The examples are from the Middlebury dataset, with a scaling factor of $\times$32.

\section{Training set up}

    We train our method with an Adam optimizer \cite{kingma2014adam} with $\beta_1=0.9$, $\beta_2=0.999$, a learning of $10^{-3}$, and weight decay parameter set to $10^{-5}$. Furthermore, we clip gradients to a maximum norm of 0.01 for stability during training.
    We find that DADA reaches convergence after 4500, 550, and 300 epochs for the datasets of Middlebury, NYUv2, and DIML respectively.
    During training, we apply data augmentation, which includes horizontal flips, random cropping, and rotating the samples by up to 15 degrees. We normalize all depth maps with their respective dataset-specific standard deviation. The Mean is not subtracted since the adjustment step assumes positive values.
    
    The feature extractor is a U-Net~\cite{ronneberger2015u} with ResNet-50~\cite{he2016deep} backbone that operates on a $\times2$ upsampled guide. Subsequently, the produced feature maps are downsampled again to the original spatial resolution. We ablate this design choice in \cref{SM:further_ablations}.

\section{Experimental set up}

    Our experiments and baselines closely follow the setup described in \cite{de2022learning} (supplementary material) and compare with mostly the same baselines.
    All methods are trained with the default settings of the Adam optimizer and a learning rate of $10^{-4}$ (except PMBA: $10^{-3}$, FDSR: $5\times10^{-4}$). We train all learned models until convergence, which is reached after 2500, 250, and 150 epochs on the datasets of Middlebury, NYUv2, and DIML respectively. We reduce the learning rate by a factor of 0.9 every 100, 10 and 6 epochs for Middlebury, NYUv2, and DIML, respectively.
    In order to limit the GPU memory consumption to a manageable level, we had to reduce the patch size for some baselines from 256$^2$ to 128$^2$ for PMBA x4 even to $64^2$. 
    For the guided filter, we use a radius of 8, and for the SD Filter, we use the following hyperparameter configuration: $\lambda=0.1$, $\sigma_g=60$, $\sigma_u=30$.
    For Pixtransform, we use the standard hyperparameters reported in their paper \cite{lutio2019guided}.
    
\section{Stability \& Statistics}

\begin{table}[t]
\centering
\footnotesize
% \vspace{-0.2cm}
 \resizebox{0.45\textwidth}{!}{%
\begin{tabular}{l| cccc}
  \multicolumn{1}{c}{Scale} & x4 & x8 & x16 & x32 \\ \midrule 
 MSE $[$cm$^2]$       & 2.52$\pm$0.04 & 5.63$\pm$0.09 & 15.6$\pm$0.10   & 47.6$\pm$0.50  \\
 MAE $[$cm$]$         & 0.11$\pm$0.00 & 0.20$\pm$0.00 & 0.47$\pm$0.00  & 1.35$\pm$0.01 \\
 MAPE $[\%]$          & 0.04$\pm$0.00 & 0.07$\pm$0.00 & 0.17$\pm$0.00  & 0.46$\pm$0.00 \\
 VV $[\%]$            & 0.06$\pm$0.00 & 0.12$\pm$0.00 & 0.30$\pm$0.00  & 0.91$\pm$0.01 \\
 EE $[\%]$            & 0.82$\pm$0.02 & 1.43$\pm$0.01 & 2.92$\pm$0.02  & 8.09$\pm$0.03 \\ \midrule 
 MSE $[$cm$^2]$       & 4.87$\pm$0.03 & 17.1$\pm$0.30 & 59.2$\pm$0.60   & 223$\pm$3.00\\
 MAE $[$cm$]$         & 0.64$\pm$0.00 & 1.33$\pm$0.01 & 2.65$\pm$0.03   & 5.76$\pm$0.02 \\
 MAPE $[\%]$          & 0.17$\pm$0.00 & 0.36$\pm$0.00 & 0.73$\pm$0.01   & 1.62$\pm$0.01 \\
 VV $[\%]$            & 0.05$\pm$0.00 & 0.18$\pm$0.00 & 0.66$\pm$0.01   & 2.40$\pm$0.03 \\
 EE $[\%]$            & 3.68$\pm$0.06 & 9.90$\pm$0.20 & 23.2$\pm$0.30   & 44.1$\pm$0.08 \\\midrule 
 MSE $[$cm$^2]$       & 1.30$\pm$0.02 & 2.87$\pm$0.06 & 7.75$\pm$0.12   & 38.6$\pm$0.80 \\
 MAE $[$cm$]$         & 0.17$\pm$0.00 & 0.27$\pm$0.00 & 0.60$\pm$0.01   & 1.90$\pm$0.02 \\
 MAPE $[\%]$          & 0.06$\pm$0.00 & 0.10$\pm$0.00 & 0.21$\pm$0.00   & 0.68$\pm$0.01 \\
 VV $[\%]$            & 0.03$\pm$0.00 & 0.07$\pm$0.00 & 0.18$\pm$0.00   & 0.85$\pm$0.01 \\
 EE $[\%]$            & 2.44$\pm$0.01 & 3.75$\pm$0.03 & 7.39$\pm$0.08   & 19.5$\pm$0.19 \\
\end{tabular}%
 }%\usepackage{adjustbox}
% \vspace{-0.1cm}
\caption{Variability of metrics on the Middlebury, NYUv2, and DIML datasets in the top, middle, and bottom sections, respectively.}
% \vspace{-0.1cm}
\label{tab:r5_SM}
\end{table}
    We also repeated the training using 4 additional (5 total) different random seeds to determine the epistemic uncertainty/stability of our method.
    We observe little variability across runs that consistently surpass previous results, see \cref{tab:r5_SM}.
    % DADA shows little variability across runs that consistently exceed previous results, see \cref{tab:r5}.
    % the baselines.

%-------------------------------------------------------------------------

\section{GPU memory consumption}
    
\begin{figure}[t]
    \centering
    
    \begin{subfigure}{0.9\linewidth}
        \centering
        \includegraphics[width=\textwidth]{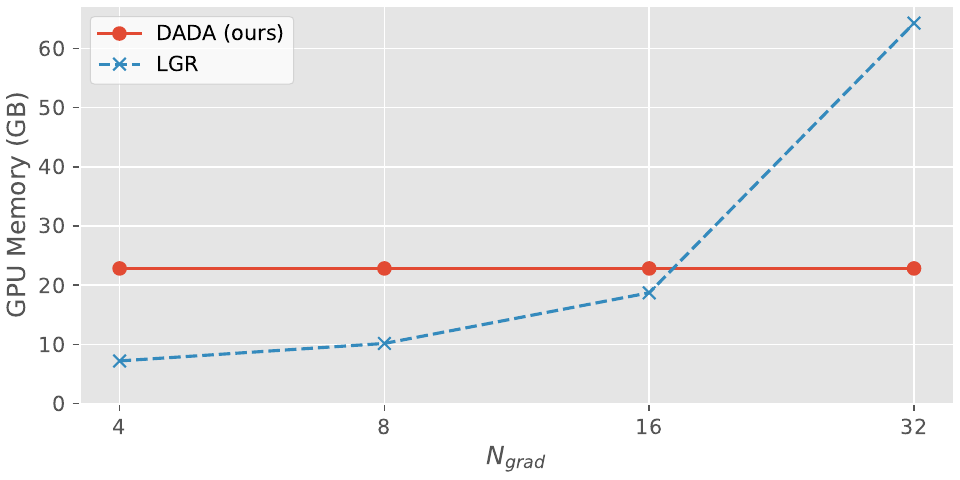}
        \caption{Memory requirement during training}
        \label{fig:ablation_mem_train}
    \end{subfigure}
    
    \begin{subfigure}{0.9\linewidth}
        \centering
        \includegraphics[width=\textwidth]{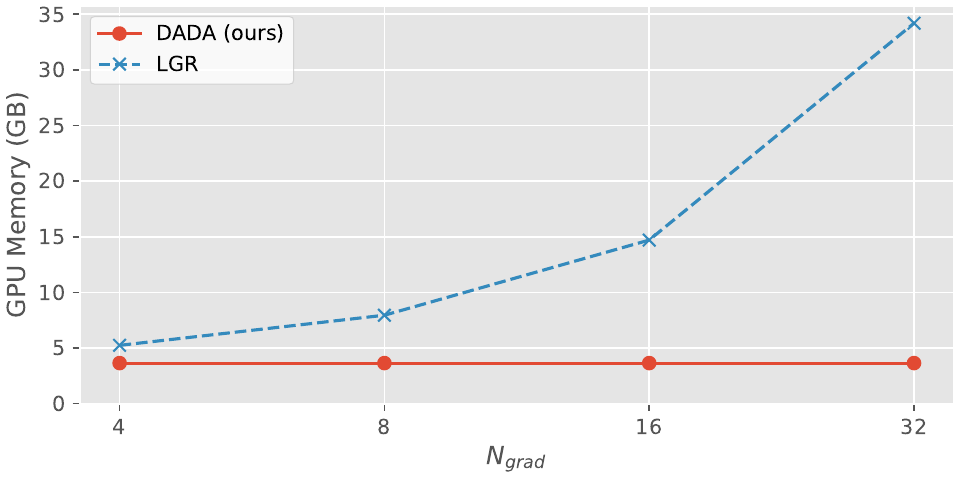}
        \caption{Memory requirement during testing}
        \label{fig:ablation_mem_test}
    \end{subfigure}
    
    \caption{Memory requirements for a batch of 8 samples for DADA and LGR. Constant memory requirements are an advantage at higher scaling factors.}
    \label{fig:ablations_mem}
\end{figure}

    In \Cref{fig:ablations_mem} we show the GPU memory consumption of our method at training time (a) and at inference (b) and contrast them to the LGR approach, which is our closest competitor and is also a hybrid method. DADA requires $\sim$23 GB at training time and $\sim$4 GB at inference time, regardless of the upsampling factor, while LGR uses up to $\sim$64 GB for training and $\sim$35 GB for testing.

\section{Inference time}

    We measure a mean inference time for the proposed method of $\sim$100~ms per sample when using a batch size of 32 on an NVIDIA GeForce RTX 3090, with all other parameters equal to the canonical ones previously described. We note that the inference time is invariant \wrt the upsampling factor. For LGR we found it to increase with the scaling factor: 50~ms, 84~ms, 240~ms, and 910~ms for $\times$4, $\times$8, $\times$16, and $\times$32, respectively. In contrast, the feedforward methods are a lot faster -- most of them are below 20~ms.

\section{Further ablations} 
\label{SM:further_ablations}

    \paragraph{Upsampling of the guide.} We explore two different ways of employing a U-Net feature extractor: Our default case where we  upsample the guide by $\times2$ and we downsample the obtained feature maps afterward and a case without up- and downsampling modifications. \cref{tab:updown_abl}, shows that this modification brings an improvement in the settings of $\times4$ to $\times$16, while a slight weakening effect is observed for $\times32$.
    Our intuition is that upsampling helps to produce sharp and accurately localized edges, which is especially beneficial for smaller scaling factors. Inversely, we speculate that for $\times32$ global context plays a bigger role, and hence, upsampling the guide decreases the receptive field of the U-Net, which in turn deteriorates the method's performance.

\begin{figure}[t]
    \centering
    
    \includegraphics[width=0.475\textwidth]{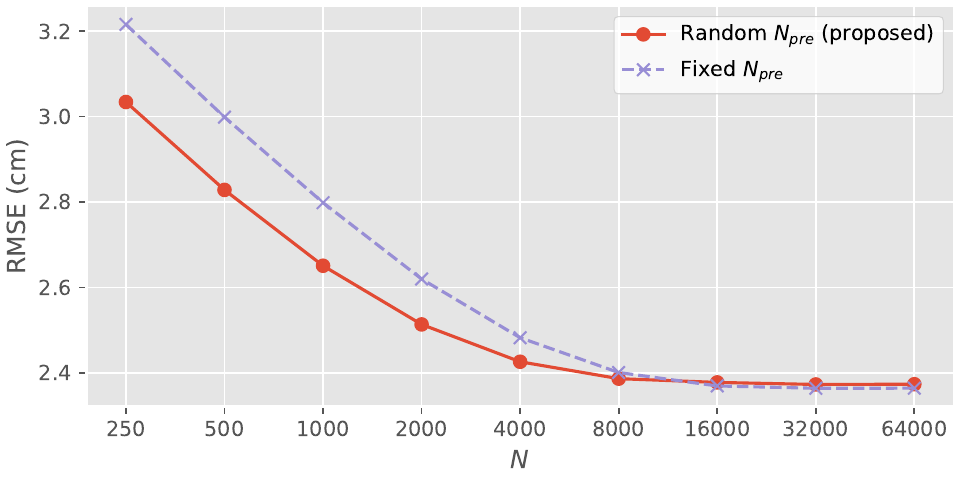}

    \caption{Randomizing the number of iterations without gradient during training speeds up convergence at inference time.}
    \label{fig:ablations_random}
\end{figure}

    \paragraph{Randomization of iterations.} In \cref{fig:ablations_random} we show the effect of randomizing the number of iterations without gradient at training time (as proposed) against choosing it to be a constant $N_{\text{grad}} = 8000$. At test time, it seems that both strategies eventually converge to an equally performant solution. However, we note that our proposed way of training leads to faster convergence in the diffusion-adjustment iterations.

    \paragraph{Feature Extractor.} We explored additional feature extractors by replacing the base case (U-Net with ResNet-50) with two more different ResNet depths (18 and 34) and  EfficientNets (B0, B1, B2). We also tested different feature extractors (FPN, DeepLabv3+) with the ResNet-50 backbone. 
    % See \cref{tab:encoders}.
    We show the results in \cref{tab:encoders_SM}. DADA is robust to changes in the exact architecture of the feature extractor.
    
    \begin{table}[t]
\centering
\footnotesize
 \resizebox{0.475\textwidth}{!}{%
\begin{tabular}{l| cccccccc}
 \multicolumn{1}{c}{} & U-Net & U-Net & \textbf{U-Net} & U-Net & U-Net & U-Net & FPN  & DL3+  \\ 
 \multicolumn{1}{c}{}  & RN18 & RN34  & \textbf{RN50}  & ENB0 & ENB1 & ENB2  & RN50 & RN50  \\  \midrule 
  % \#layers & & & & & & & \\ \midrule  
 MSE          & 5.69  & 5.57 & 5.63  & 6.85 & 6.77 & 6.68 & 7.09 & 7.77 \\
 MAE          & 0.20  & 0.20 & 0.20  & 0.24 & 0.24 & 0.24 & 0.27 & 0.30 \\
 
\end{tabular}%
 }%\usepackage{adjustbox}
% \vspace{-0.1cm}
\caption{Middlebury, scale x8. DADA is invariant to the encoder depth and fairly invariant to the choice of model architecture.}
% \vspace{-0.1cm}
\label{tab:encoders_SM}
\end{table}

\begin{table}[t]
    \centering
    \resizebox{0.475\textwidth}{!}{%
    \begin{tabular}{l | cccc}
    % \toprule
     & $\times4$  & $\times8$  & $\times16$  & $\times32$  \\
     & MSE / MAE & MSE / MAE & MSE / MAE & MSE / MAE  \\ \midrule \midrule
    \multicolumn{5}{c}{\textbf{Middlebury}}  \\ \midrule 
    \textbf{Base} & \textbf{2.58} / \textbf{0.11} & \textbf{5.63} / \textbf{0.20} & \textbf{16.3} / \textbf{0.48} & 50.6 / 1.38 \\ 
    \quad No up-/down & 2.88 / 0.12 & 5.92 / 0.22 & 18.3 / 0.55 & \textbf{49.9} / \textbf{1.36} \\
    %\bottomrule

    \midrule \midrule 

     \multicolumn{5}{c}{\textbf{NYUv2}}  \\ \midrule 
    \textbf{Base} & \textbf{4.83} / \textbf{0.64} & \textbf{16.6} / \textbf{1.30} & \textbf{59.0} / \textbf{2.64} & 228 / 5.81 \\ 
    \quad No up-/down  & 6.12 / 0.72 & 18.7 / 1.40 & 60.7 / 2.72 & \textbf{207} / \textbf{5.56} \\%\bottomrule

    \midrule \midrule 

    \multicolumn{5}{c}{\textbf{DIML}}  \\ \midrule 
    \textbf{DADA} & \textbf{1.33} / \textbf{0.17} & \textbf{2.93} / \textbf{0.28} & \textbf{7.61} / \textbf{0.59} & 39.8 / 1.92 \\
    \quad No up-/down & 1.59 / 0.18 & 3.28 / 0.30 & 8.56 / 0.64 & \textbf{36.8} / \textbf{1.82} \\%\bottomrule
    \end{tabular}%
    }
    \caption{Ablation: method as proposed \vs method which skips upsampling before and downsampling after the feature extractor. Errors are in cm$^2$ (MSE) and in cm (MAE). Up-/downsampling appears to have a positive effect for lower scales and a neutral or slightly negative effect for very large scales.}
    \label{tab:updown_abl}
\end{table}

\section{Additional metrics} \label{add_metrics}

We concur that other metrics can provide additional insight into the method's performance and facilitate future comparisons.
We computed the mean absolute percentage error (MAPE), value errors (VE), and edge errors (EE) as described in \cite{he2021}. The advantage of DADA \wrt current methods remains clear, see \cref{tab:new_metrics_SM}.

\begin{table}[t]
\centering
\footnotesize
% \vspace{-0.075cm}
 \resizebox{0.45\textwidth}{!}{%
\begin{tabular}{l| ccc}
 \multicolumn{1}{c}{} & \textbf{Middlebury} & \textbf{NYUv2} & \textbf{DIML}  \\ 
 \multicolumn{1}{c}{in $[\%]$} &       MAPE/VE/EE & MAPE/VE/EE & MAPE/VE/EE  \\ \midrule  
 MSG          &  0.8 / 1.7 / 14.7 &  1.9 / 3.0 / 49.8 &  0.9 / 1.2 / 24.8 \\
 DKN          &  1.2 / 2.7 / 21.9 &  2.4 / 4.4 / 55.8 &  1.4 / 2.2 / 32.3 \\
 FDKN         &  1.2 / 2.6 / 21.7 &  2.4 / 4.5 / 55.7 &  1.4 / 2.2 / 34.3 \\
 PMBA         &  1.9 / 4.1 / 35.7 &  2.7 / 5.2 / 57.4 &  0.8 / 1.2 / 25.0 \\
 FDSR         &  1.6 / 3.7 / 21.6 &  3.4 / 7.2 / 63.4 &  1.6 / 2.5 / 36.3 \\
 LGR          &  0.6 / 1.2 / 12.8 &  1.8 / 2.7 / 49.0 &  0.9 / 1.3 / 26.7 \\
 DADA         &  \textbf{0.5} / \textbf{0.9} / \textbf{8.1} &  \textbf{1.6} / \textbf{2.4} / \textbf{44.3} &  \textbf{0.7} / \textbf{0.9} / \textbf{19.7}
\end{tabular}%
 }%\usepackage{adjustbox}
 % \vspace{-0.1cm}
\caption{Middlebury, scale x32, all numbers are in [\%].}
\vspace{-0.2cm}
\label{tab:new_metrics_SM}
\end{table}

\section{Main results as curves}

\begin{figure*}[htbp]
    \centering
    % \vspace{10cm}
    
    \begin{subfigure}{0.31\linewidth}
        \centering
        \includegraphics[width=\textwidth]{figures/curve_rmse_middlebury_sup.pdf}
        \caption{Middlebury}
        \label{fig:rmse_m_sup}
    \end{subfigure}
    \hfill
    \begin{subfigure}{0.31\linewidth}
        \centering
        \includegraphics[width=\textwidth]{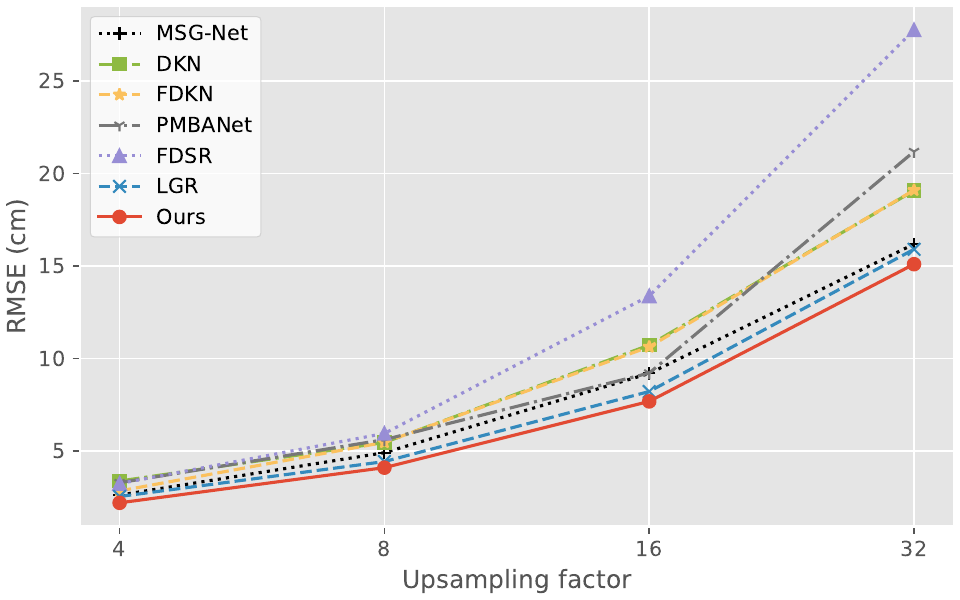}
        \caption{NYUv2}
        \label{fig:rmse_n_sup}
    \end{subfigure}
    \hfill
    \begin{subfigure}{0.31\linewidth}
        \centering
        \includegraphics[width=\textwidth]{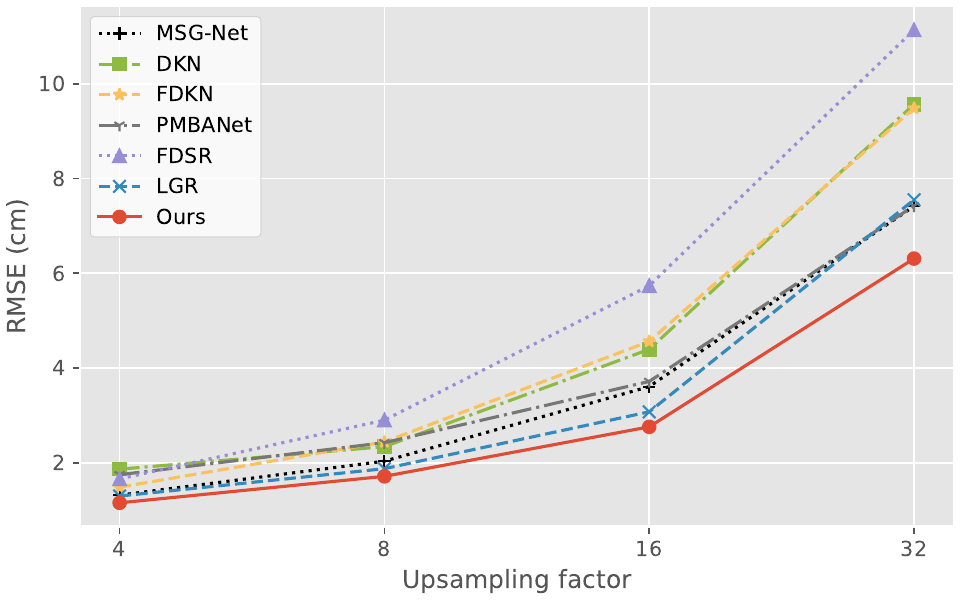}
        \caption{DIML}
        \label{fig:rmse_d_sup}
    \end{subfigure}
    
    \begin{subfigure}{0.31\linewidth}
        \centering
        \includegraphics[width=\textwidth]{figures/curve_rmse_middlebury_unsup.pdf}
        \caption{Middlebury}
        \label{fig:rmse_m_unsup}
    \end{subfigure}
    \hfill
    \begin{subfigure}{0.31\linewidth}
        \centering
        \includegraphics[width=\textwidth]{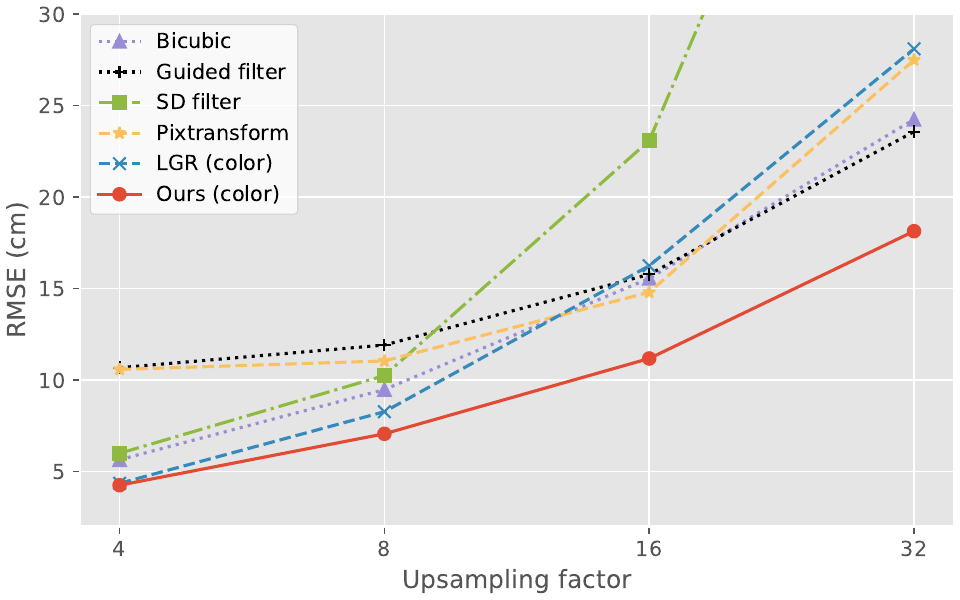}
        \caption{NYUv2}
        \label{fig:rmse_n_unsup}
    \end{subfigure}
    \hfill
    \begin{subfigure}{0.31\linewidth}
        \centering
        \includegraphics[width=\textwidth]{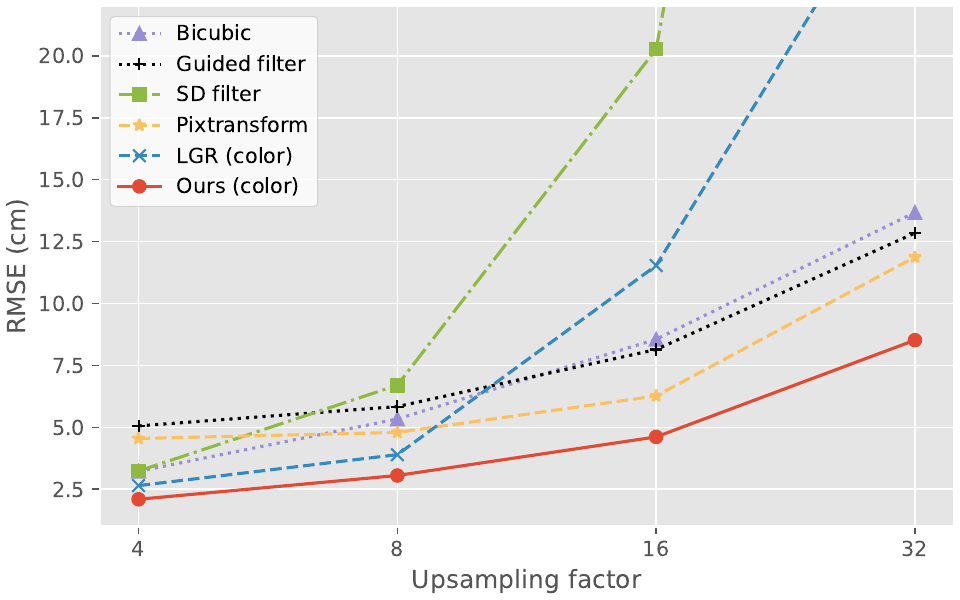}
        \caption{DIML}
        \label{fig:rmse_d_unsup}
    \end{subfigure}
    
    \caption{Curves with numbers from Tables 1 and 2 in main paper for better interpretability of results. RMSE is used instead of MSE for improved visualization across scales.}
    \label{fig:all_curves}
\end{figure*}

    For improved interpretation, the RMSE numbers from Tables 1 and 2 from the main paper are here also displayed as curves. These plots can be seen in \Cref{fig:all_curves}.

\section{Intuition behind the adjustment step}
    
\begin{figure}[t]
        \centering
        \vspace{-0.3cm}
        \includegraphics[width=0.475\textwidth]{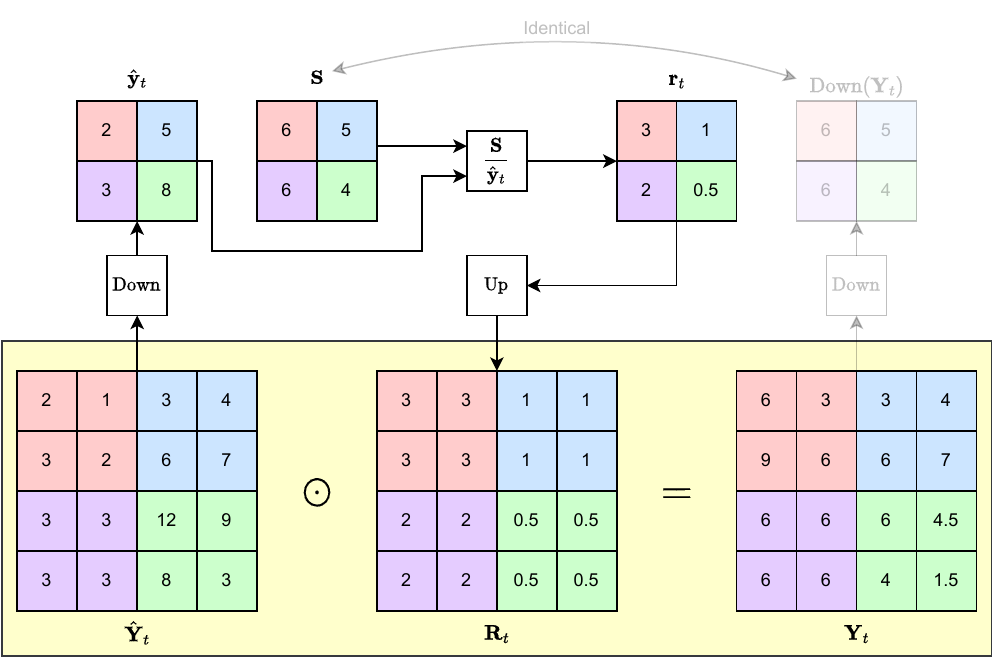}
        \vspace{-0.3cm}
        \caption{Numerical example of the adjustment step. 
        }
        \label{fig:adjustment_intuition}
        \vspace{-0.1cm}
    \end{figure}

    The values of $\mathbf{r}_{t}$ and $\mathbf{R}_{t}$ are computed precisely so that $\text{down}(\mathbf{Y}_{t})$ matches $\mathbf{S}$. The values in $\mathbf{r}_{t}$ are per-patch adjustment coefficients. A numerical example of these operations is shown in \cref{fig:adjustment_intuition} to help convey the intuition behind this operation.
    
\section{Qualitative results}
    
    We provide additional results from for all learned methods, the color version of LGR, and Pixtransform -- the strongest unsupervised competitor -- for all three datasets and all four scaling factors in \Cref{fig:sup_viz_1}, \Cref{fig:sup_viz_2}, and \Cref{fig:sup_viz_3}.
    
    The plots how that for smaller upsampling factors, all learned methods perform relatively well, while for larger upsampling factors, the advantages of DADA become more apparent: Shapes are represented more accurately and edges are sharper.

\newpage
\FloatBarrier

\setlength{\tabcolsep}{1pt} 

\newpage
\FloatBarrier
\newcommand\MIDv{400}
\newcommand\MIDe{6}
\newcommand\MIDs{17}
\newcommand\MIDt{41}

\setlength{\tabcolsep}{1pt}
\newlength{\imww}
\setlength{\imww}{0.10\textwidth}

\begin{sidewaystable}[p]
\vspace{10cm}
  \centering
  \footnotesize
\begin{tabular}{cccccc|ccccccccc}
% \parbox[t]{2mm}{\multirow{2}{*}{\rotatebox[origin=c]{90}{\textbf{Middlebury}}}} &
%\rotatebox[origin=c]{90}{$\times$16} &\
\multirow{2}{*}[0.35cm]{\rotatebox{270}{\textbf{$\times$4}}} &
\includegraphics[width=\imww]{images/Middlebury/idx_\MIDv/guide_\MIDv.png} &
\includegraphics[width=\imww]{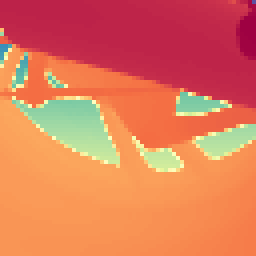} &
\includegraphics[width=\imww]{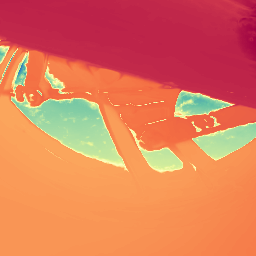} &
\includegraphics[width=\imww]{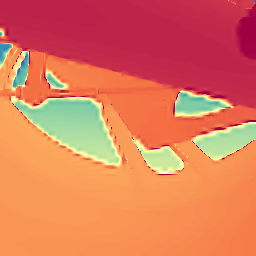} &
\includegraphics[width=\imww]{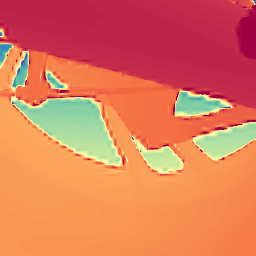} & 
\includegraphics[width=\imww]{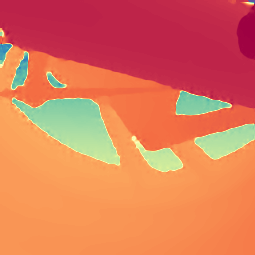} &
\includegraphics[width=\imww]{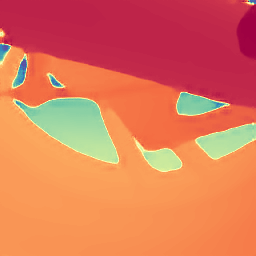} &
\includegraphics[width=\imww]{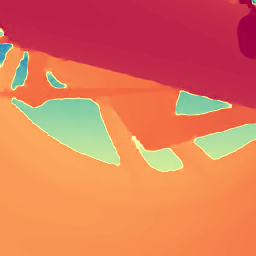} &
\includegraphics[width=\imww]{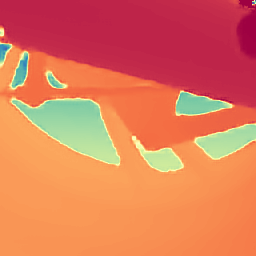} &
\includegraphics[width=\imww]{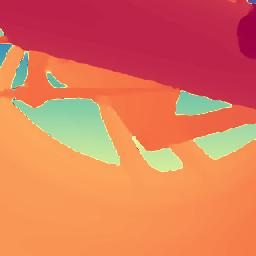} & 
\includegraphics[width=\imww]{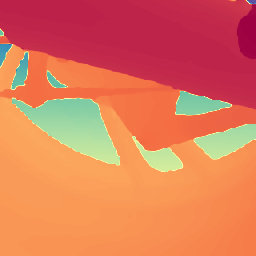} &
\includegraphics[width=\imww]{images/Middlebury/idx_\MIDv/gt_idx_\MIDv.png} \\
 & &
 &
\includegraphics[width=\imww]{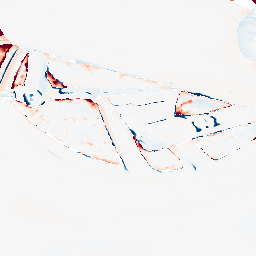} &
\includegraphics[width=\imww]{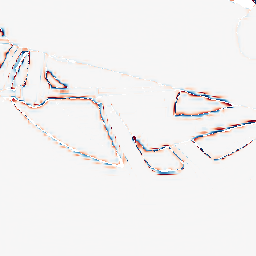} &
\includegraphics[width=\imww]{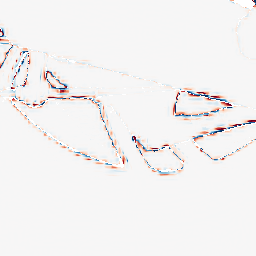} & 
\includegraphics[width=\imww]{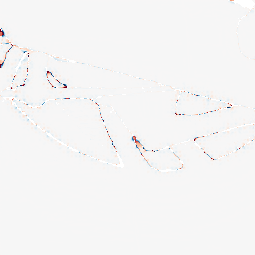} &
\includegraphics[width=\imww]{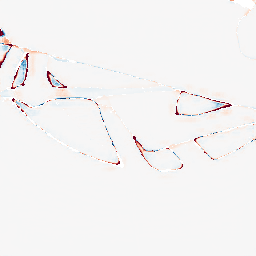} &
\includegraphics[width=\imww]{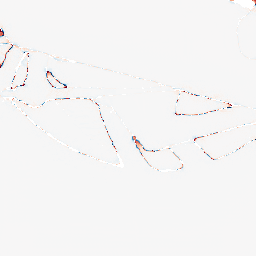} &
\includegraphics[width=\imww]{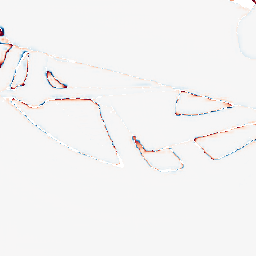} &
\includegraphics[width=\imww]{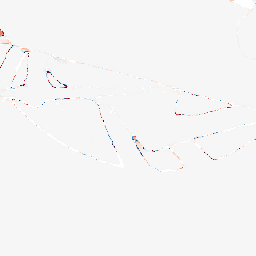} &
\includegraphics[width=\imww]{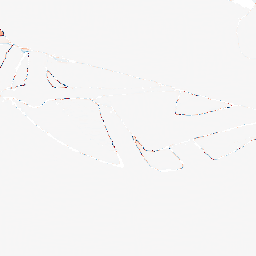} & 
\\
% \parbox[t]{2mm}{\multirow{2}{*}{\rotatebox[origin=c]{90}{\textbf{Middlebury}}}} &
%\rotatebox[origin=c]{90}{$\times$8} &
\multirow{2}{*}[0.35cm]{\rotatebox{270}{\textbf{$\times$8}}} &
\includegraphics[width=\imww]{images/Middlebury/idx_\MIDe/guide_\MIDe.png} &
\includegraphics[width=\imww]{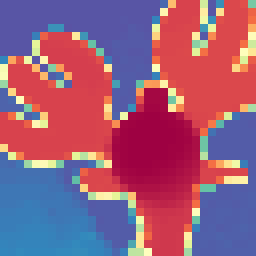} &
\includegraphics[width=\imww]{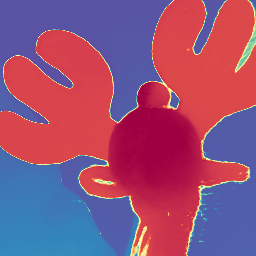} &
\includegraphics[width=\imww]{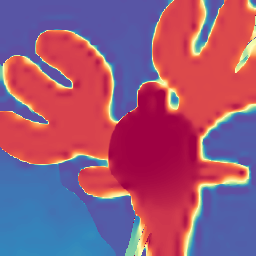} &
\includegraphics[width=\imww]{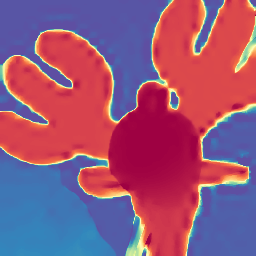} & 
\includegraphics[width=\imww]{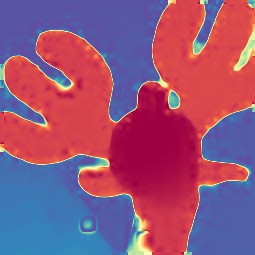} &
\includegraphics[width=\imww]{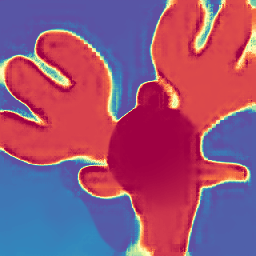} &
\includegraphics[width=\imww]{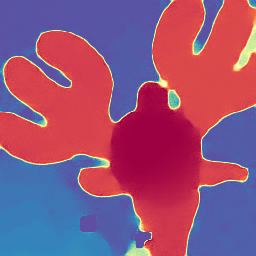} &
\includegraphics[width=\imww]{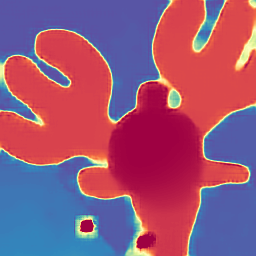} &
\includegraphics[width=\imww]{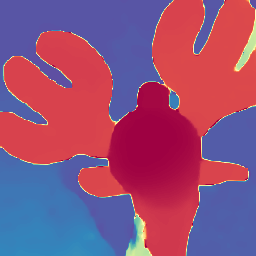} & 
\includegraphics[width=\imww]{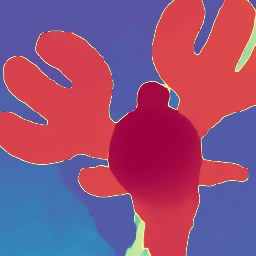} &
\includegraphics[width=\imww]{images/Middlebury/idx_\MIDe/gt_idx_\MIDe.png} \\ 
 & &
 &
\includegraphics[width=\imww]{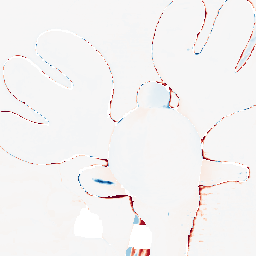} &
\includegraphics[width=\imww]{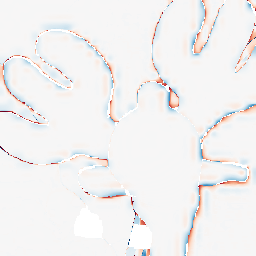} &
\includegraphics[width=\imww]{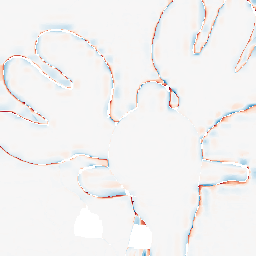} & 
\includegraphics[width=\imww]{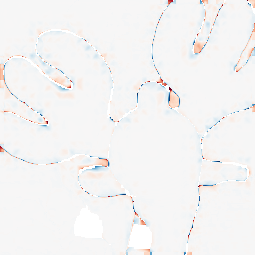} &
\includegraphics[width=\imww]{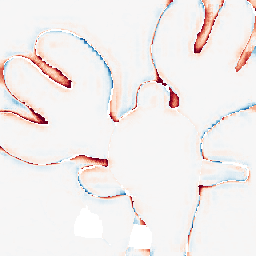} &
\includegraphics[width=\imww]{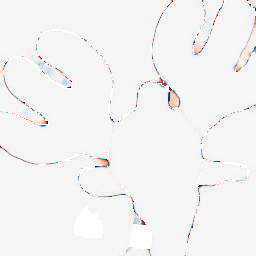} &
\includegraphics[width=\imww]{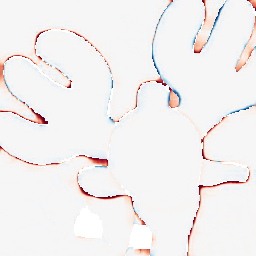} &
\includegraphics[width=\imww]{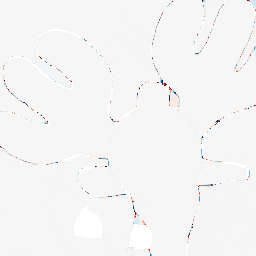} &
\includegraphics[width=\imww]{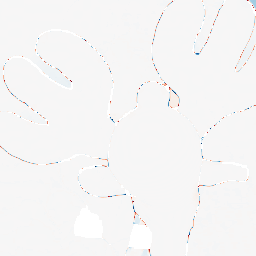} & 
\\
% \parbox[t]{2mm}{\multirow{2}{*}{\rotatebox[origin=c]{90}{\textbf{Middlebury}}}} &
%\rotatebox[origin=c]{90}{$\times$16} &
\multirow{2}{*}[0.35cm]{\rotatebox{270}{\textbf{$\times$16}}} &
\includegraphics[width=\imww]{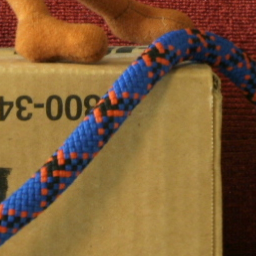} &
\includegraphics[width=\imww]{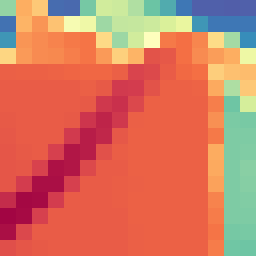} &
\includegraphics[width=\imww]{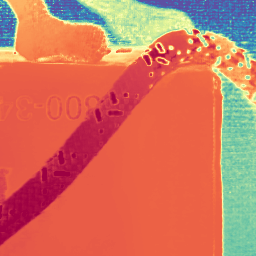} &
\includegraphics[width=\imww]{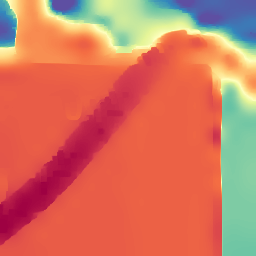} &
\includegraphics[width=\imww]{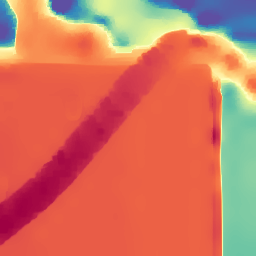} & 
\includegraphics[width=\imww]{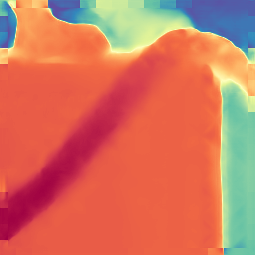} &
\includegraphics[width=\imww]{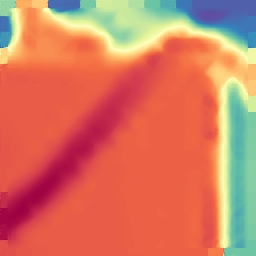} &
\includegraphics[width=\imww]{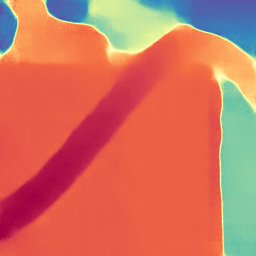} &
\includegraphics[width=\imww]{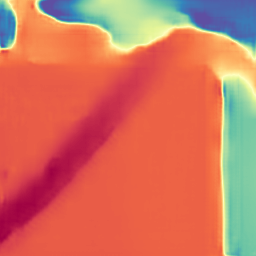} &
\includegraphics[width=\imww]{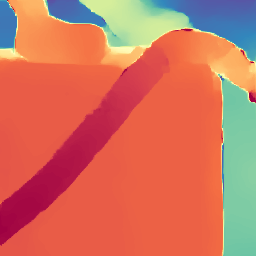} & 
\includegraphics[width=\imww]{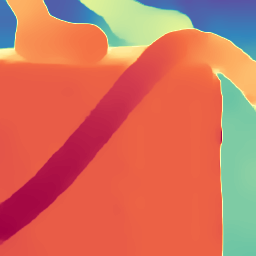} &
\includegraphics[width=\imww]{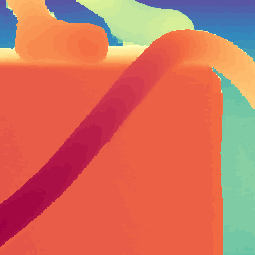} & 
\\
 & &
 &
\includegraphics[width=\imww]{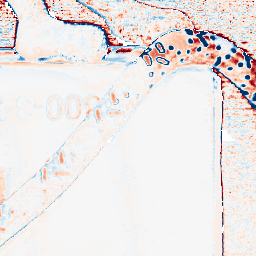} &
\includegraphics[width=\imww]{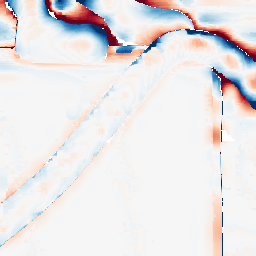} &
\includegraphics[width=\imww]{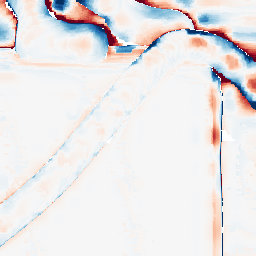} & 
\includegraphics[width=\imww]{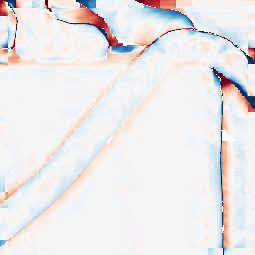} &
\includegraphics[width=\imww]{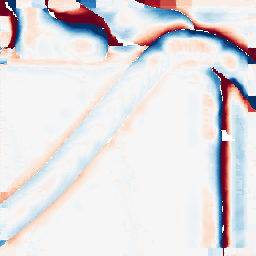} &
\includegraphics[width=\imww]{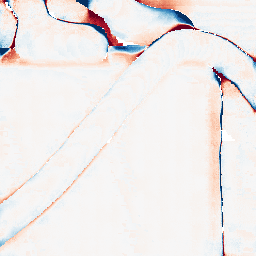} &
\includegraphics[width=\imww]{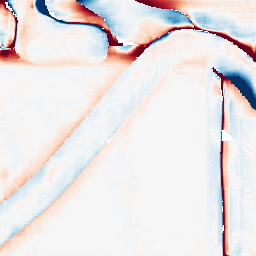} &
\includegraphics[width=\imww]{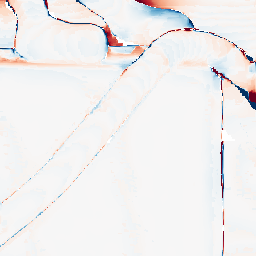} &
\includegraphics[width=\imww]{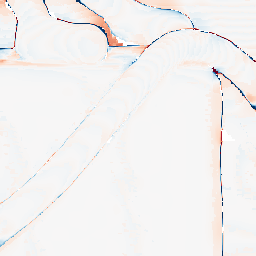} & 
\\
\multirow{2}{*}[0.35cm]{\rotatebox{270}{\textbf{$\times$32}}} &
\includegraphics[width=\imww]{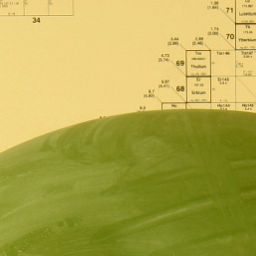} &
\includegraphics[width=\imww]{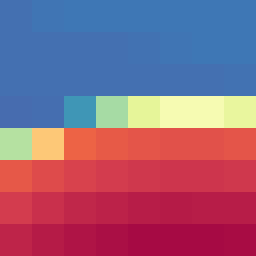} &
\includegraphics[width=\imww]{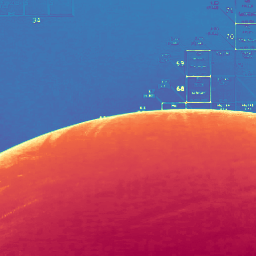} &
\includegraphics[width=\imww]{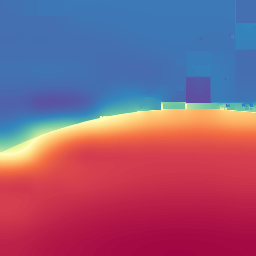} &
\includegraphics[width=\imww]{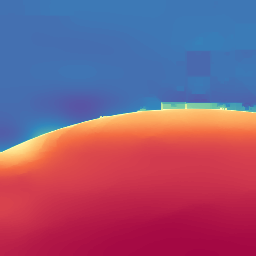} &
\includegraphics[width=\imww]{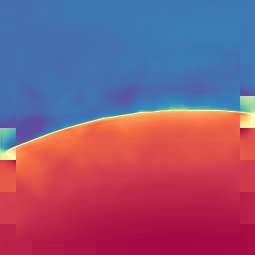} &
\includegraphics[width=\imww]{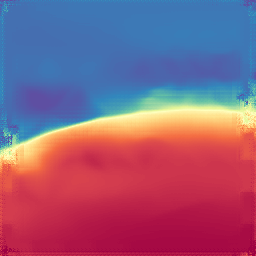} &
\includegraphics[width=\imww]{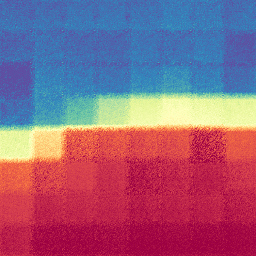} &
\includegraphics[width=\imww]{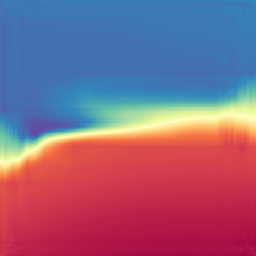} &
\includegraphics[width=\imww]{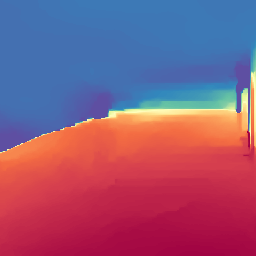} &
\includegraphics[width=\imww]{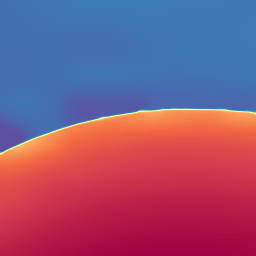} & 
\includegraphics[width=\imww]{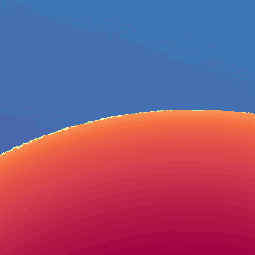} &
 \\
 & &
&
\includegraphics[width=\imww]{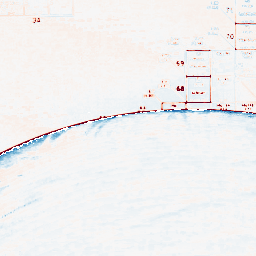} &
\includegraphics[width=\imww]{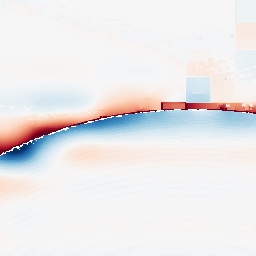} &
\includegraphics[width=\imww]{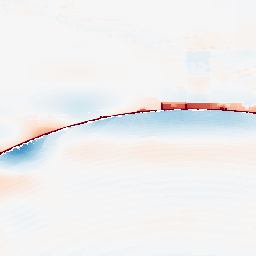} &
\includegraphics[width=\imww]{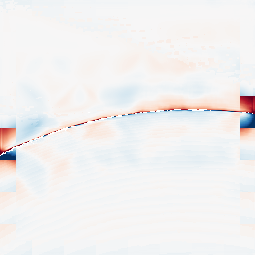} &
\includegraphics[width=\imww]{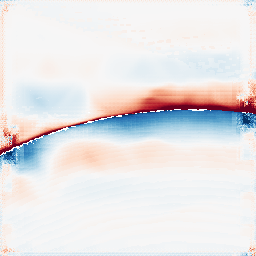} &
\includegraphics[width=\imww]{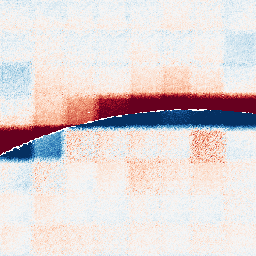} &
\includegraphics[width=\imww]{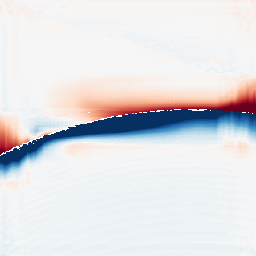} &
\includegraphics[width=\imww]{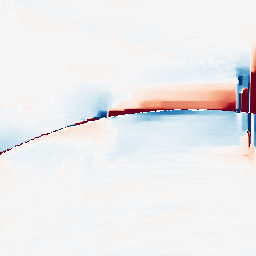} &
\includegraphics[width=\imww]{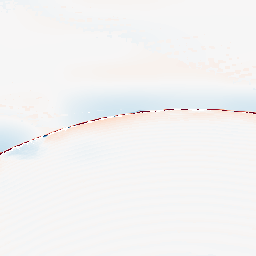} &  \\ 
& Guide & Source & PixT & LGR$^\dag$ & DADA$^\dag$ & MSG & FDKN & PMBA & FDSR & LGR & DADA & GT
\end{tabular}
\vspace{-0.5em}
\caption{Predictions and error maps for different guided super-resolution methods on the Middlebury dataset. Blue denotes under-estimated depth, red denotes over-estimation. All plots in a row have the same color scale. The last two columns juxtapose our predictions and the ground truth. }
\label{fig:sup_viz_1}
\end{sidewaystable}

\newpage
\FloatBarrier
\newcommand\Nv{323}
\newcommand\Ne{97}
\newcommand\Ns{3}
\newcommand\Nt{5}

\setlength{\tabcolsep}{1pt}
\newlength{\imN}
\setlength{\imN}{0.10\textwidth}

\begin{sidewaystable}[p]
\vspace{10cm}
  \centering
  \footnotesize
\begin{tabular}{cccccc|ccccccccc}
% \parbox[t]{2mm}{\multirow{2}{*}{\rotatebox[origin=c]{90}{\textbf{NYUv2}}}} &
%\rotatebox[origin=c]{90}{$\times$16} &\
\multirow{2}{*}[0.35cm]{\rotatebox{270}{\textbf{$\times$4}}} &
\includegraphics[width=\imN]{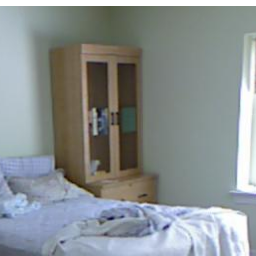} &
\includegraphics[width=\imN]{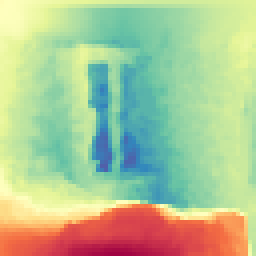} &
\includegraphics[width=\imN]{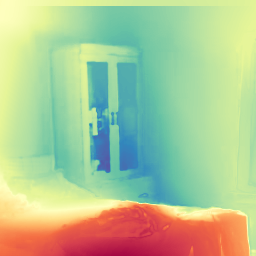} &
\includegraphics[width=\imN]{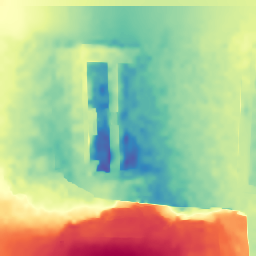} &
\includegraphics[width=\imN]{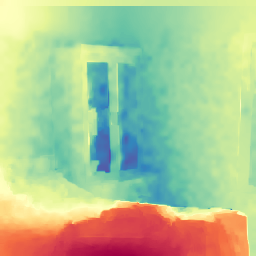} & 
\includegraphics[width=\imN]{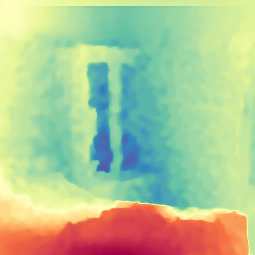} &
\includegraphics[width=\imN]{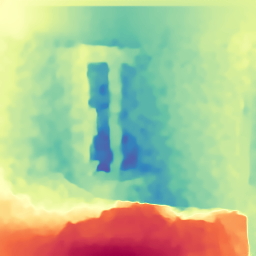} &
\includegraphics[width=\imN]{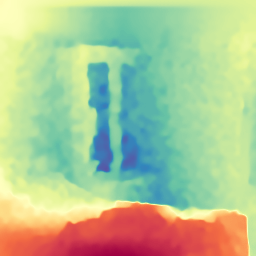} &
\includegraphics[width=\imN]{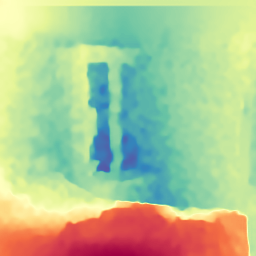} &
\includegraphics[width=\imN]{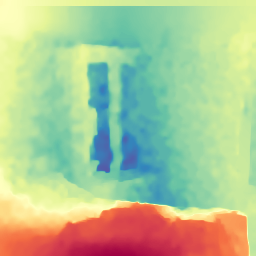} & 
\includegraphics[width=\imN]{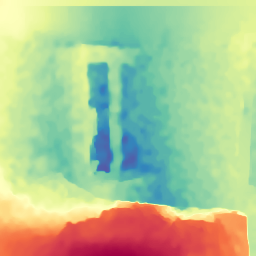} &
\includegraphics[width=\imN]{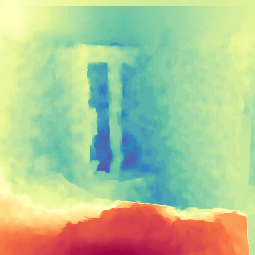} \\
 & &
 &
\includegraphics[width=\imN]{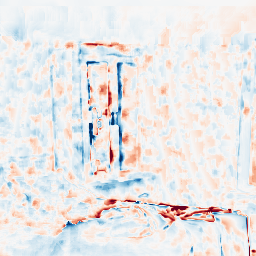} &
\includegraphics[width=\imN]{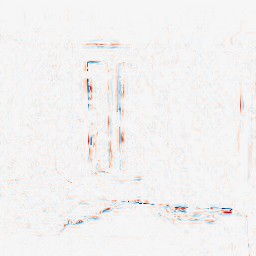} &
\includegraphics[width=\imN]{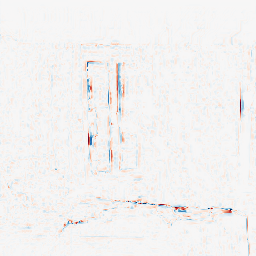} & 
\includegraphics[width=\imN]{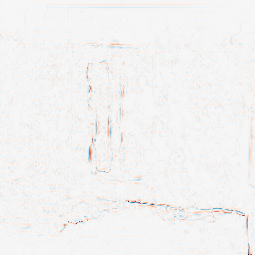} &
\includegraphics[width=\imN]{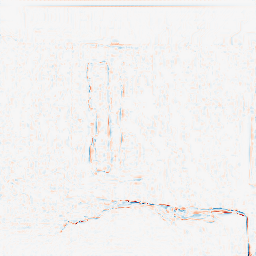} &
\includegraphics[width=\imN]{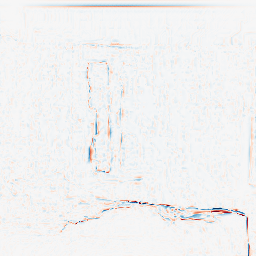} &
\includegraphics[width=\imN]{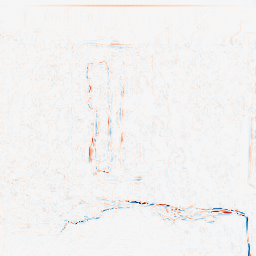} &
\includegraphics[width=\imN]{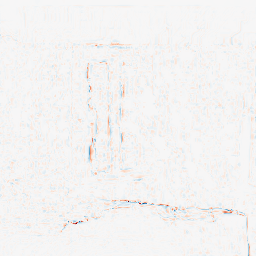} &
\includegraphics[width=\imN]{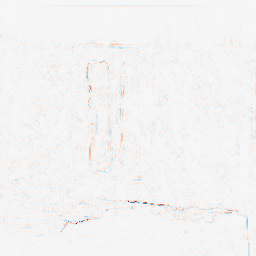} & 
\\
% \parbox[t]{2mm}{\multirow{2}{*}{\rotatebox[origin=c]{90}{\textbf{NYUv2}}}} &
%\rotatebox[origin=c]{90}{$\times$8} &
\multirow{2}{*}[0.35cm]{\rotatebox{270}{\textbf{$\times$8}}} &
\includegraphics[width=\imN]{images/NYUv2/idx_\Ne/guide_\Ne.png} &
\includegraphics[width=\imN]{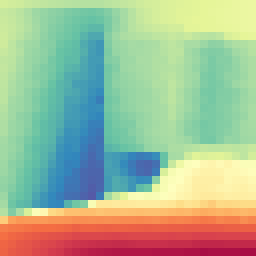} &
\includegraphics[width=\imN]{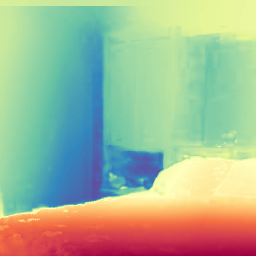} &
\includegraphics[width=\imN]{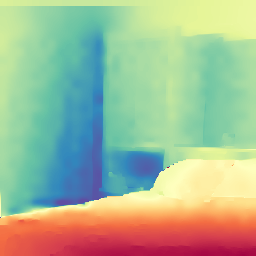} &
\includegraphics[width=\imN]{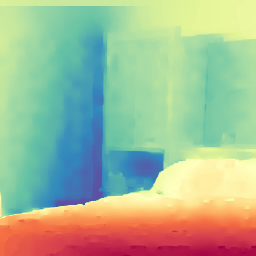} & 
\includegraphics[width=\imN]{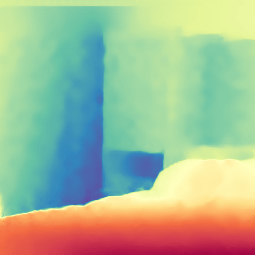} &
\includegraphics[width=\imN]{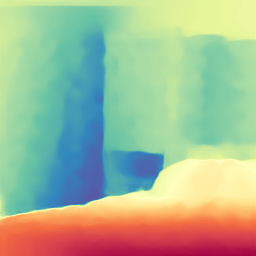} &
\includegraphics[width=\imN]{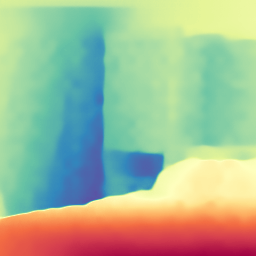} &
\includegraphics[width=\imN]{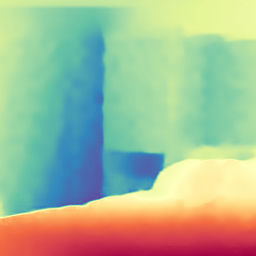} &
\includegraphics[width=\imN]{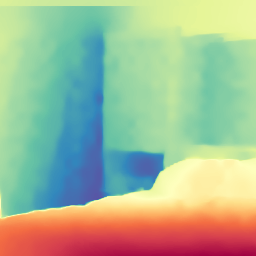} & 
\includegraphics[width=\imN]{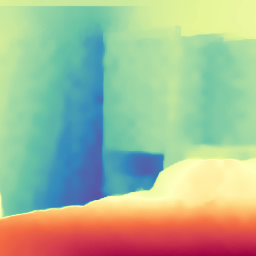} &
\includegraphics[width=\imN]{images/NYUv2/idx_\Ne/gt_idx_\Ne.png} \\ 
 & &
 &
\includegraphics[width=\imN]{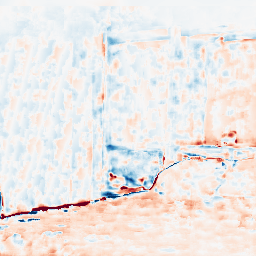} &
\includegraphics[width=\imN]{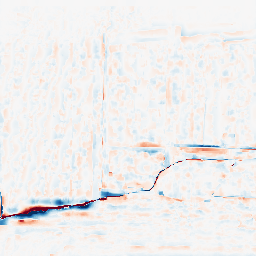} &
\includegraphics[width=\imN]{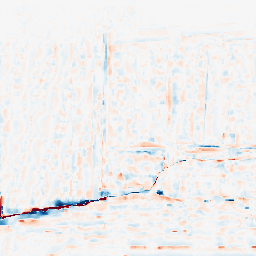} & 
\includegraphics[width=\imN]{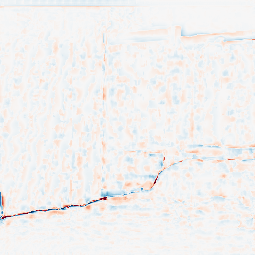} &
\includegraphics[width=\imN]{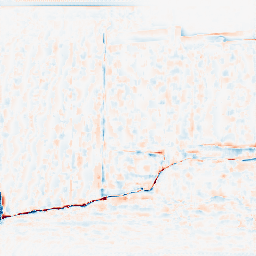} &
\includegraphics[width=\imN]{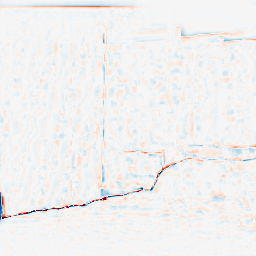} &
\includegraphics[width=\imN]{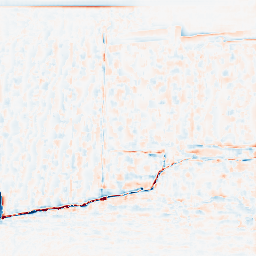} &
\includegraphics[width=\imN]{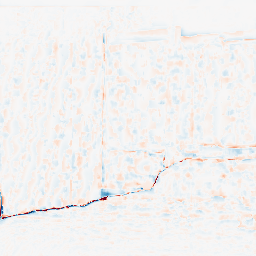} &
\includegraphics[width=\imN]{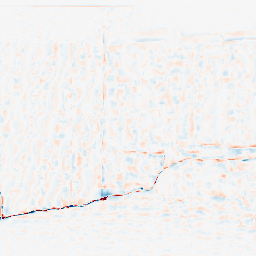} & 
\\
% \parbox[t]{2mm}{\multirow{2}{*}{\rotatebox[origin=c]{90}{\textbf{NYUv2}}}} &
%\rotatebox[origin=c]{90}{$\times$16} &
\multirow{2}{*}[0.35cm]{\rotatebox{270}{\textbf{$\times$16}}} &
\includegraphics[width=\imN]{images/NYUv2/idx_\Ns/guide_\Ns.png} &
\includegraphics[width=\imN]{images/NYUv2/idx_\Ns/source_idx_\Ns_scale_16.png} &
\includegraphics[width=\imN]{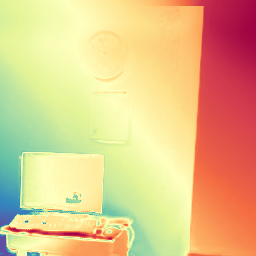} &
\includegraphics[width=\imN]{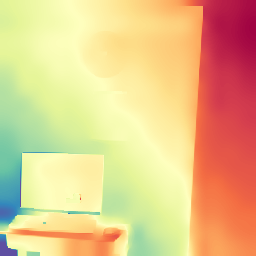} &
\includegraphics[width=\imN]{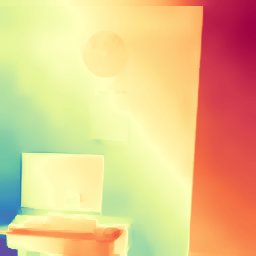} & 
\includegraphics[width=\imN]{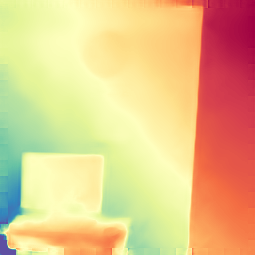} &
\includegraphics[width=\imN]{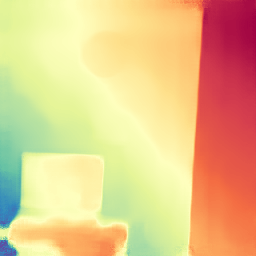} &
\includegraphics[width=\imN]{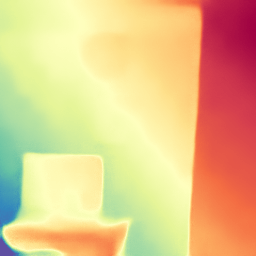} &
\includegraphics[width=\imN]{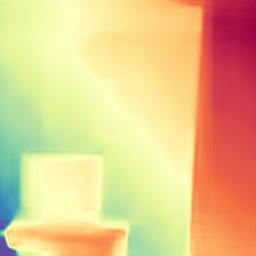} &
\includegraphics[width=\imN]{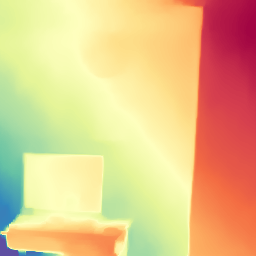} & 
\includegraphics[width=\imN]{images/NYUv2/idx_\Ns/DADA_idx_\Ns_scale_16.png} &
\includegraphics[width=\imN]{images/NYUv2/idx_\Ns/gt_idx_\Ns.png} & 
\\
 & &
 &
\includegraphics[width=\imN]{mydiffs/NYUv2/idx_\Ns/pixtransform_idx_\Ns_scale_16.png} &
\includegraphics[width=\imN]{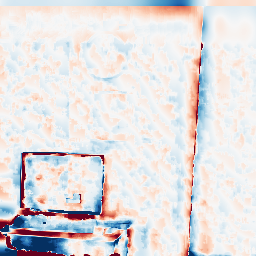} &
\includegraphics[width=\imN]{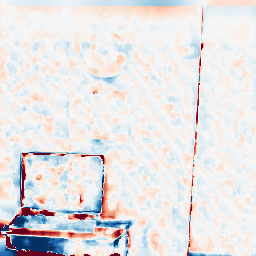} & 
\includegraphics[width=\imN]{mydiffs/NYUv2/idx_\Ns/MSGNet_idx_\Ns_scale_16.png} &
\includegraphics[width=\imN]{mydiffs/NYUv2/idx_\Ns/fdkn_idx_\Ns_scale_16.png} &
\includegraphics[width=\imN]{mydiffs/NYUv2/idx_\Ns/pmba_full_idx_\Ns_scale_16.png} &
\includegraphics[width=\imN]{mydiffs/NYUv2/idx_\Ns/FDSR_theirs_idx_\Ns_scale_16.png} &
\includegraphics[width=\imN]{mydiffs/NYUv2/idx_\Ns/LGR_idx_\Ns_scale_16.png} &
\includegraphics[width=\imN]{mydiffs/NYUv2/idx_\Ns/DADA_idx_\Ns_scale_16.png} & 
\\
\multirow{2}{*}[0.35cm]{\rotatebox{270}{\textbf{$\times$32}}} &
\includegraphics[width=\imN]{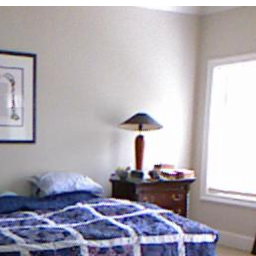} &
\includegraphics[width=\imN]{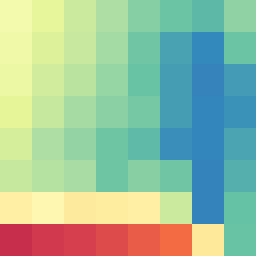} &
\includegraphics[width=\imN]{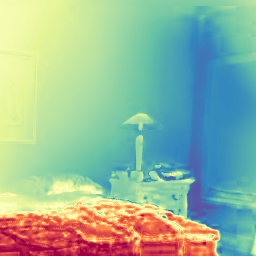} &
\includegraphics[width=\imN]{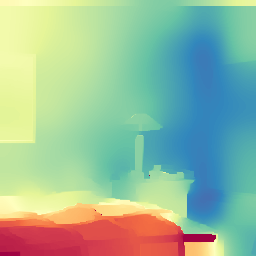} &
\includegraphics[width=\imN]{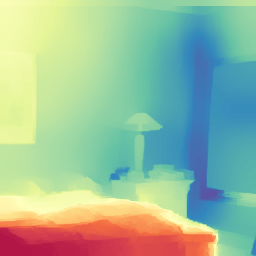} &
\includegraphics[width=\imN]{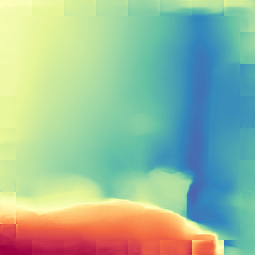} &
\includegraphics[width=\imN]{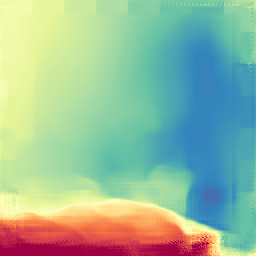} &
\includegraphics[width=\imN]{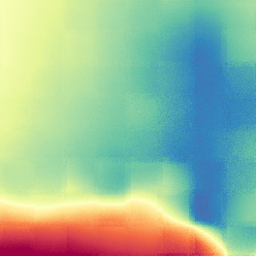} &
\includegraphics[width=\imN]{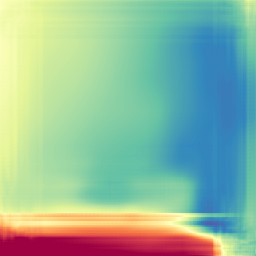} &
\includegraphics[width=\imN]{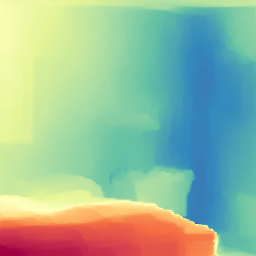} &
\includegraphics[width=\imN]{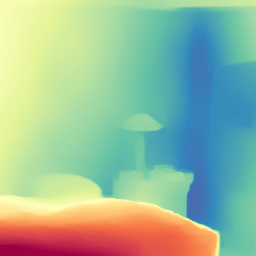} & 
\includegraphics[width=\imN]{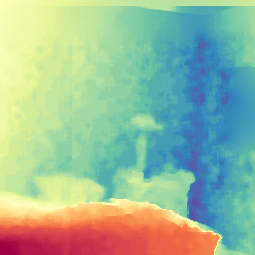} &
 \\
 & &
&
\includegraphics[width=\imN]{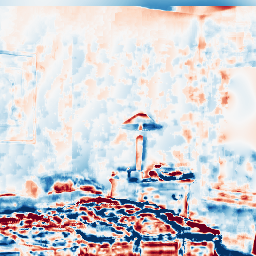} &
\includegraphics[width=\imN]{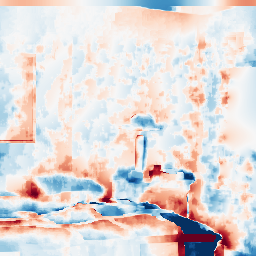} &
\includegraphics[width=\imN]{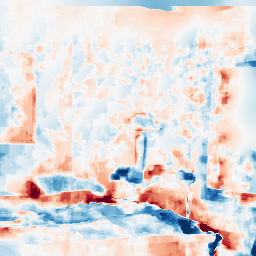} &
\includegraphics[width=\imN]{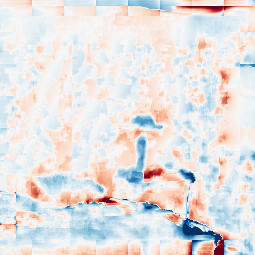} &
\includegraphics[width=\imN]{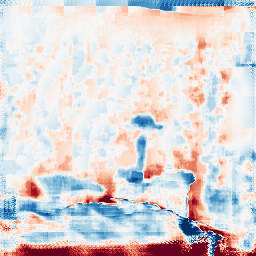} &
\includegraphics[width=\imN]{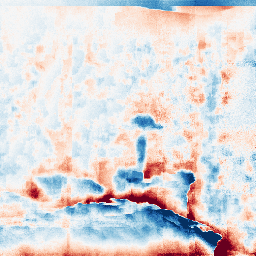} &
\includegraphics[width=\imN]{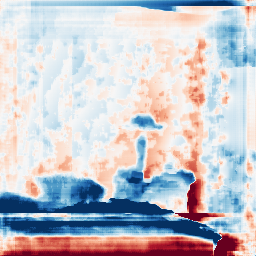} &
\includegraphics[width=\imN]{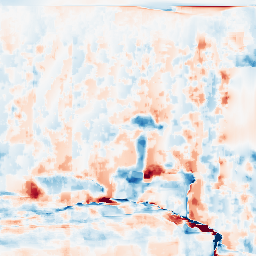} &
\includegraphics[width=\imN]{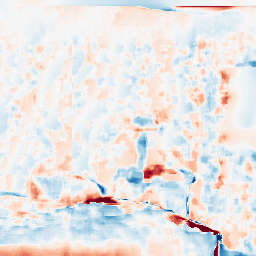} &  \\ 
& Guide & Source & PixT & LGR$^\dag$ & DADA$^\dag$ & MSG & FDKN & PMBA & FDSR & LGR & DADA & GT
\end{tabular}
\vspace{-0.5em}
\caption{Predictions and error maps for different guided super-resolution methods on the NYUv2 dataset. Blue denotes under-estimated depth, red denotes over-estimation. All plots in a row have the same color scale. The last two columns juxtapose our predictions and the ground truth.  }
\label{fig:sup_viz_2}
\end{sidewaystable}

\newpage
\FloatBarrier
\newcommand\Dv{200} % 0
\newcommand\De{1} % 200
\newcommand\Ds{4755} % 4755
\newcommand\Dt{4291}

\setlength{\tabcolsep}{1pt}
\newlength{\imD}
\setlength{\imD}{0.10\textwidth}

\begin{sidewaystable}[htbp]
\vspace{10cm}
  \centering
  \footnotesize
\begin{tabular}{cccccc|ccccccccc}
% \parbox[t]{2mm}{\multirow{2}{*}{\rotatebox[origin=c]{90}{\textbf{DIML}}}} &
%\rotatebox[origin=c]{90}{$\times$16} &\
\multirow{2}{*}[0.35cm]{\rotatebox{270}{\textbf{$\times$4}}} &
\includegraphics[width=\imD]{images/DIML/idx_\Dv/guide_\Dv.png} &
\includegraphics[width=\imD]{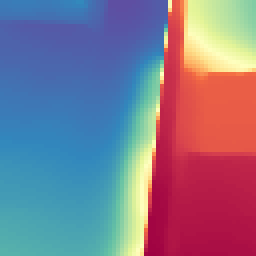} &
\includegraphics[width=\imD]{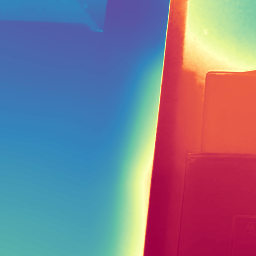} &
\includegraphics[width=\imD]{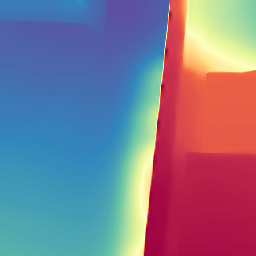} &
\includegraphics[width=\imD]{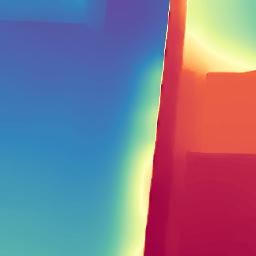} & 
\includegraphics[width=\imD]{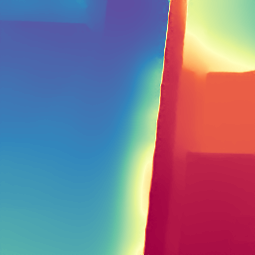} &
\includegraphics[width=\imD]{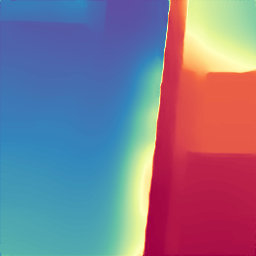} &
\includegraphics[width=\imD]{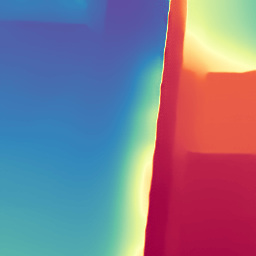} &
\includegraphics[width=\imD]{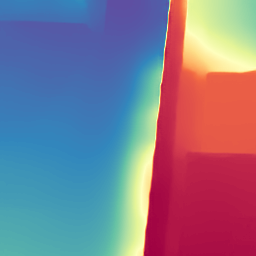} &
\includegraphics[width=\imD]{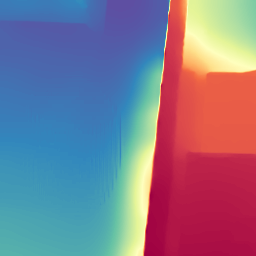} & 
\includegraphics[width=\imD]{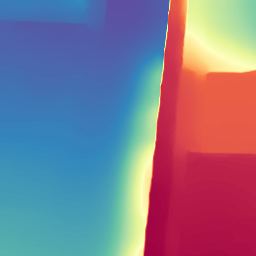} &
\includegraphics[width=\imD]{images/DIML/idx_\Dv/gt_idx_\Dv.png} \\
 & &
 &
\includegraphics[width=\imD]{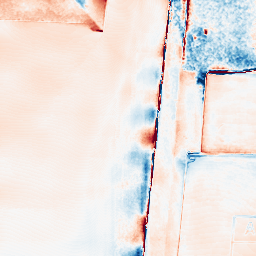} &
\includegraphics[width=\imD]{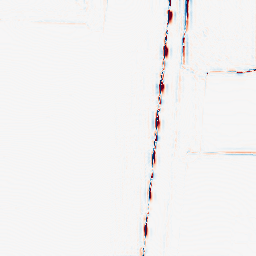} &
\includegraphics[width=\imD]{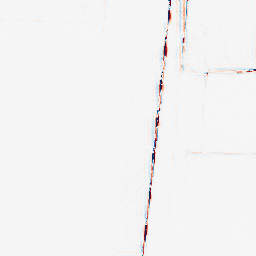} & 
\includegraphics[width=\imD]{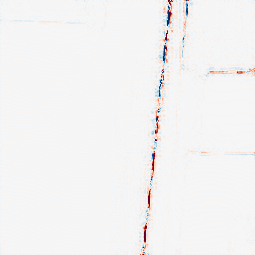} &
\includegraphics[width=\imD]{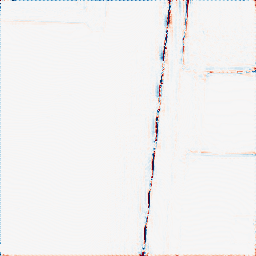} &
\includegraphics[width=\imD]{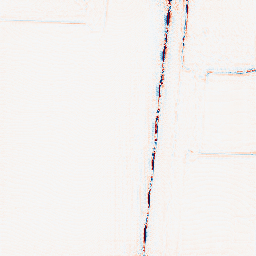} &
\includegraphics[width=\imD]{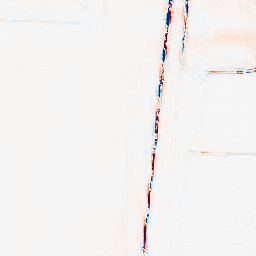} &
\includegraphics[width=\imD]{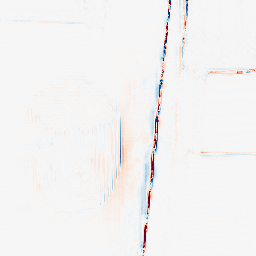} &
\includegraphics[width=\imD]{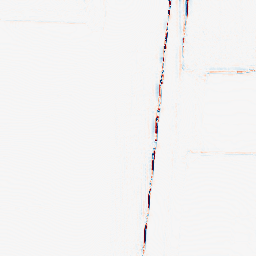} & 
\\
% \parbox[t]{2mm}{\multirow{2}{*}{\rotatebox[origin=c]{90}{\textbf{DIML}}}} &
%\rotatebox[origin=c]{90}{$\times$8} &
\multirow{2}{*}[0.35cm]{\rotatebox{270}{\textbf{$\times$8}}} &
\includegraphics[width=\imD]{images/DIML/idx_\De/guide_\De.png} &
\includegraphics[width=\imD]{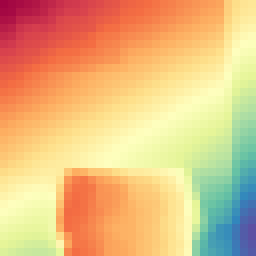} &
\includegraphics[width=\imD]{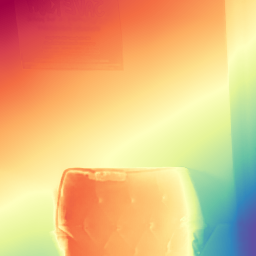} &
\includegraphics[width=\imD]{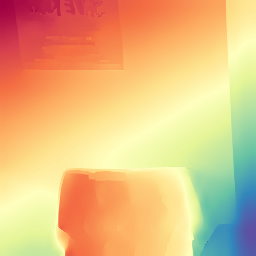} &
\includegraphics[width=\imD]{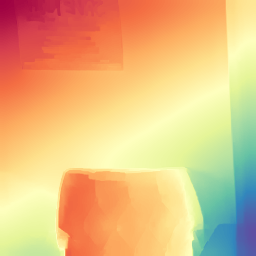} & 
\includegraphics[width=\imD]{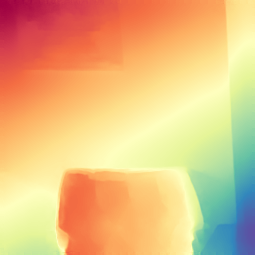} &
\includegraphics[width=\imD]{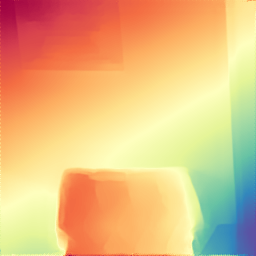} &
\includegraphics[width=\imD]{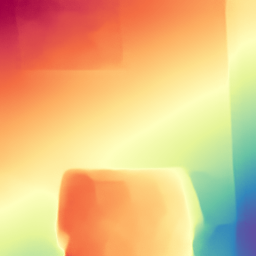} &
\includegraphics[width=\imD]{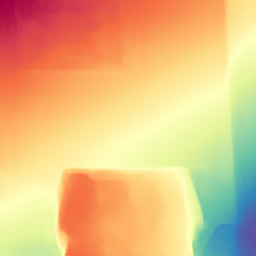} &
\includegraphics[width=\imD]{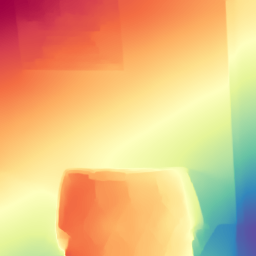} & 
\includegraphics[width=\imD]{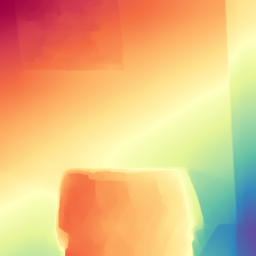} &
\includegraphics[width=\imD]{images/DIML/idx_\De/gt_idx_\De.png} \\ 
 & &
 &
\includegraphics[width=\imD]{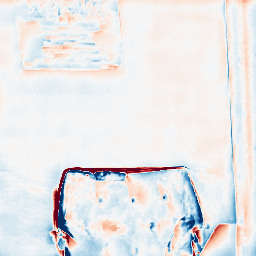} &
\includegraphics[width=\imD]{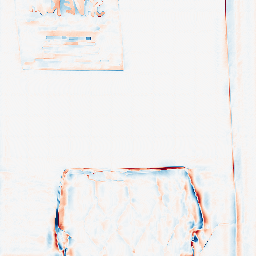} &
\includegraphics[width=\imD]{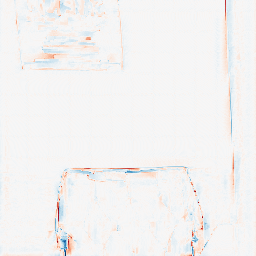} & 
\includegraphics[width=\imD]{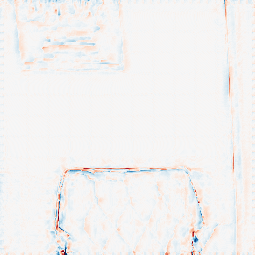} &
\includegraphics[width=\imD]{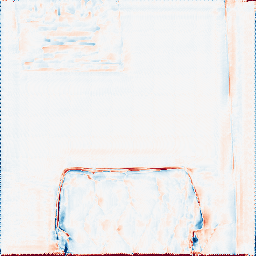} &
\includegraphics[width=\imD]{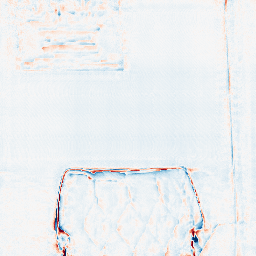} &
\includegraphics[width=\imD]{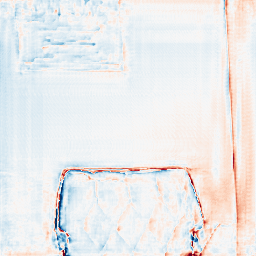} &
\includegraphics[width=\imD]{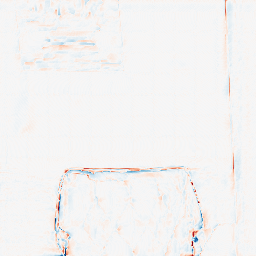} &
\includegraphics[width=\imD]{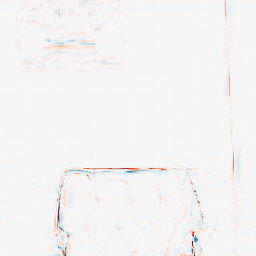} & 
\\
% \parbox[t]{2mm}{\multirow{2}{*}{\rotatebox[origin=c]{90}{\textbf{DIML}}}} &
%\rotatebox[origin=c]{90}{$\times$16} &
\multirow{2}{*}[0.35cm]{\rotatebox{270}{\textbf{$\times$16}}} &
\includegraphics[width=\imD]{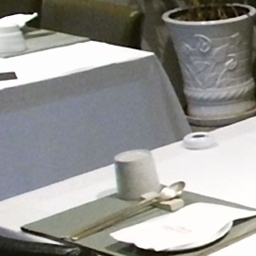} &
\includegraphics[width=\imD]{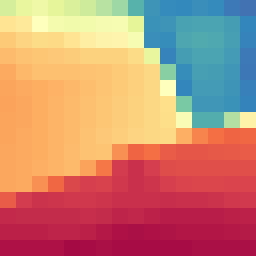} &
\includegraphics[width=\imD]{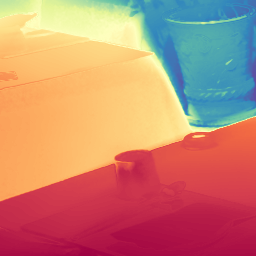} &
\includegraphics[width=\imD]{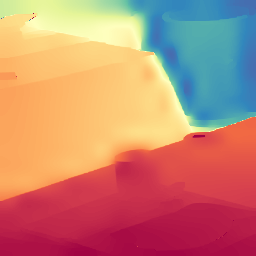} &
\includegraphics[width=\imD]{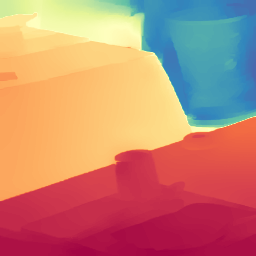} & 
\includegraphics[width=\imD]{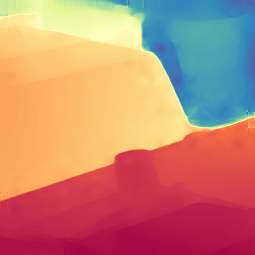} &
\includegraphics[width=\imD]{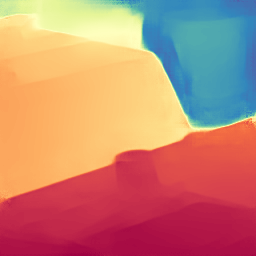} &
\includegraphics[width=\imD]{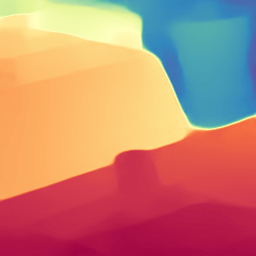} &
\includegraphics[width=\imD]{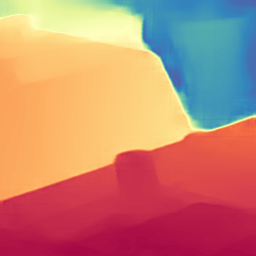} &
\includegraphics[width=\imD]{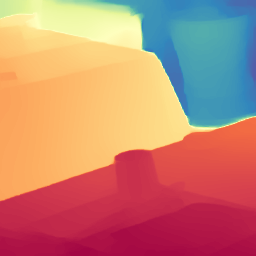} & 
\includegraphics[width=\imD]{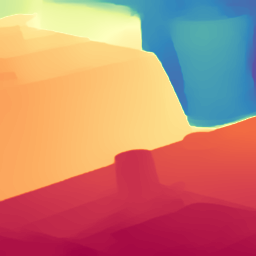} &
\includegraphics[width=\imD]{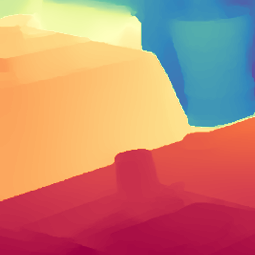} & 
\\
 & &
 &
\includegraphics[width=\imD]{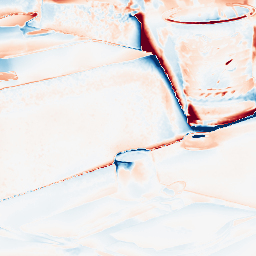} &
\includegraphics[width=\imD]{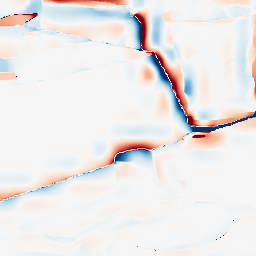} &
\includegraphics[width=\imD]{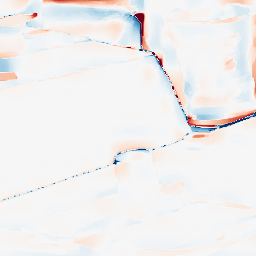} & 
\includegraphics[width=\imD]{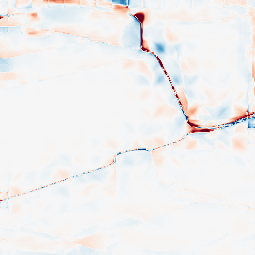} &
\includegraphics[width=\imD]{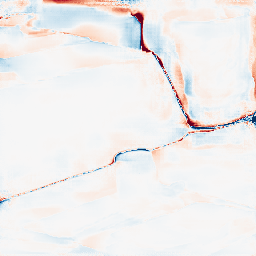} &
\includegraphics[width=\imD]{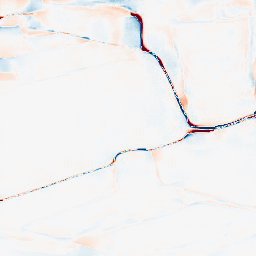} &
\includegraphics[width=\imD]{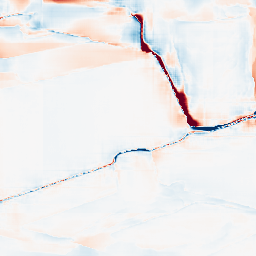} &
\includegraphics[width=\imD]{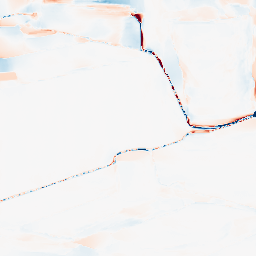} &
\includegraphics[width=\imD]{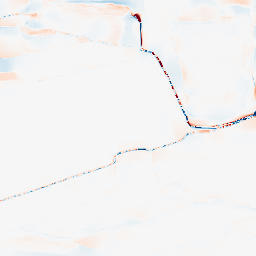} & 
\\
\multirow{2}{*}[0.35cm]{\rotatebox{270}{\textbf{$\times$32}}} &
\includegraphics[width=\imD]{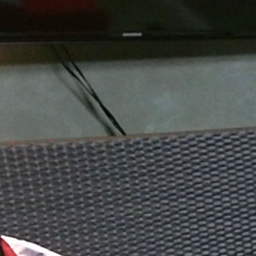} &
\includegraphics[width=\imD]{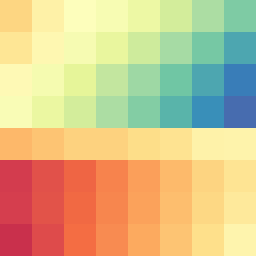} &
\includegraphics[width=\imD]{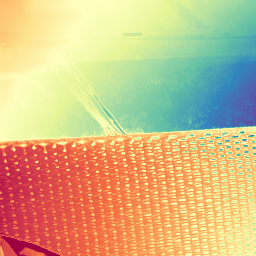} &
\includegraphics[width=\imD]{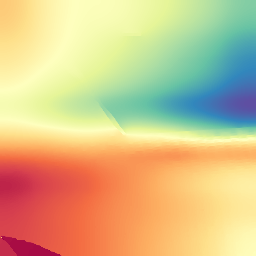} &
\includegraphics[width=\imD]{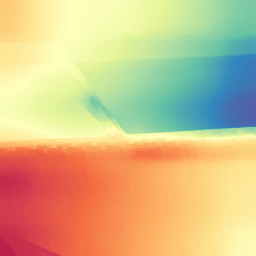} &
\includegraphics[width=\imD]{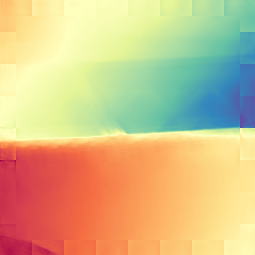} &
\includegraphics[width=\imD]{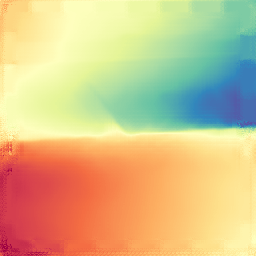} &
\includegraphics[width=\imD]{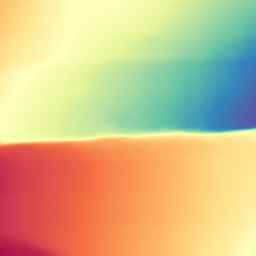} &
\includegraphics[width=\imD]{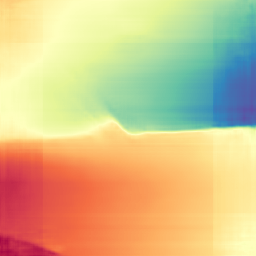} &
\includegraphics[width=\imD]{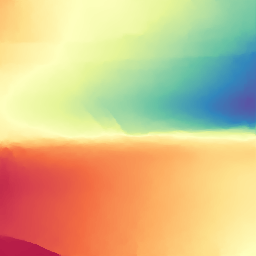} &
\includegraphics[width=\imD]{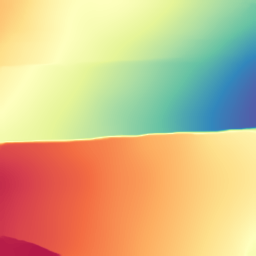} & 
\includegraphics[width=\imD]{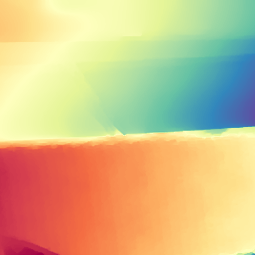} &
 \\
 & &
&
\includegraphics[width=\imD]{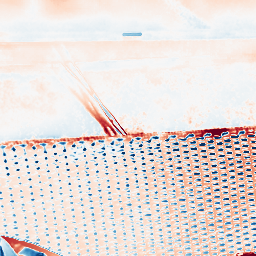} &
\includegraphics[width=\imD]{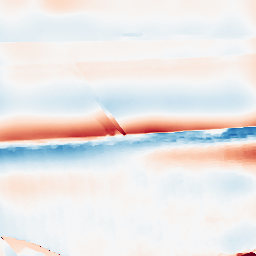} &
\includegraphics[width=\imD]{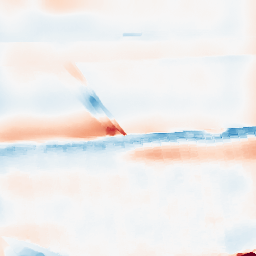} &
\includegraphics[width=\imD]{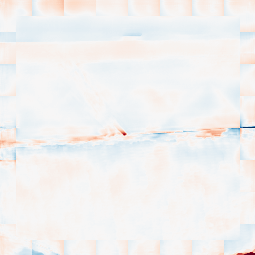} &
\includegraphics[width=\imD]{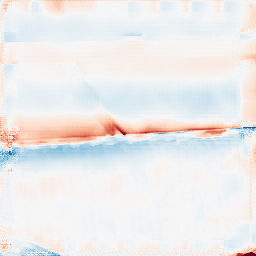} &
\includegraphics[width=\imD]{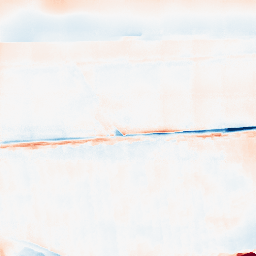} &
\includegraphics[width=\imD]{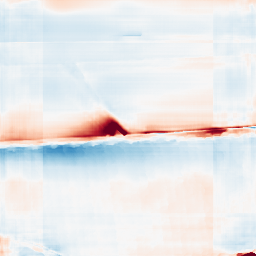} &
\includegraphics[width=\imD]{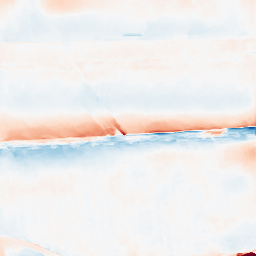} &
\includegraphics[width=\imD]{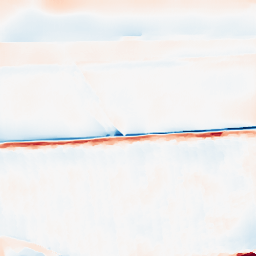} &  \\ 
& Guide & Source  & PixT & LGR$^\dag$ & DADA$^\dag$ & MSG & FDKN & PMBA & FDSR & LGR & DADA & GT
\end{tabular}
\vspace{-0.5em}
\caption{Predictions and error maps for different guided super-resolution methods on the DIML dataset. Blue denotes under-estimated depth, red denotes over-estimation. All plots in a row have the same color scale. The last two columns juxtapose our predictions and the ground truth.  }
\label{fig:sup_viz_3}
\end{sidewaystable}

\newpage
\FloatBarrier

\end{document}